\documentclass[journal,twocolumn,10pt,compsoc]{IEEEtran}

\usepackage{times}
\usepackage{epsfig}
\usepackage{graphicx}
\usepackage{amsmath}
\usepackage{amssymb}

\usepackage{caption}
\usepackage[dvipsnames]{xcolor}
\usepackage{mathtools}

\usepackage{courier}
\usepackage{makecell}
\usepackage{color, colortbl}
\definecolor{Gray}{gray}{0.5}
\usepackage{wrapfig}
\usepackage{float}
\usepackage{bbm}
\usepackage{pifont}
\usepackage{courier}
\usepackage{multirow}
\usepackage{amsfonts}
\usepackage{breakcites}
\usepackage{booktabs}
\usepackage{savesym}
\savesymbol{checkmark}
\usepackage{dingbat}
\usepackage{color}
\usepackage{url}

\usepackage{colortbl}
\usepackage{comment}
\usepackage{subfigure}

\makeatletter
\def\hlinewd#1{%
\noalign{\ifnum0=`}\fi\hrule \@height #1 \futurelet
\reserved@a\@xhline}

\newcommand{\cmark}{\ding{51}}
\newcommand{\xmark}{\ding{55}}

\definecolor{shcolor}{rgb}{0,0,1}

\definecolor{srcolor}{rgb}{1,0,0}

\definecolor{mygray}{gray}{0.9}

\ifCLASSOPTIONcompsoc
  \usepackage[nocompress]{cite}
\else
  \usepackage{cite}
\fi

\begin{document}

\title{CATs++: Boosting Cost Aggregation with Convolutions and Transformers}

\author{Seokju Cho*,~\IEEEmembership{Student Member,~IEEE,}
Sunghwan Hong*,~\IEEEmembership{Student Member,~IEEE,}~\thanks{* Equal contribution} \\
and Seungryong Kim$^{\dagger}$,~\IEEEmembership{Member,~IEEE}~\thanks{$\dagger$ Corresponding author}
\IEEEcompsocitemizethanks{
\IEEEcompsocthanksitem S. Cho, S. Hong, and S. Kim are with Korea University, Seoul, Korea.
}
}

\markboth{}%
{Shell \MakeLowercase{\textit{et al.}}: Bare Demo of IEEEtran.cls for Computer Society Journals}

\IEEEtitleabstractindextext{%
\begin{abstract}

Cost aggregation is a process in image matching tasks that aims to disambiguate the noisy matching scores. Existing methods generally tackle this by hand-crafted or CNN-based methods, which either lack robustness to severe deformations or inherit the limitation of CNNs that fail to discriminate incorrect matches due to limited receptive fields and inadaptability. In this paper, we introduce Cost Aggregation with Transformers (CATs) to tackle this by exploring global consensus among initial correlation map with the help of some architectural designs that allow us to benefit from global receptive fields of self-attention mechanism. To this end, we include appearance affinity modeling, which helps to disambiguate the noisy initial correlation maps. Furthermore, we introduce some techniques, including multi-level aggregation to exploit rich semantics prevalent at different feature levels and swapping self-attention to obtain reciprocal matching scores to act as a regularization. Although CATs can attain competitive performance, it may face some limitations, \textit{i.e.}, high computational costs, which may restrict its applicability only at limited resolution and hurt performance. To overcome this, we propose CATs++, an extension of CATs. Concretely, we introduce early convolutions prior to cost aggregation with a transformer to control the number of tokens and inject some convolutional inductive bias, then propose a novel transformer architecture for both efficient and effective cost aggregation, which results in apparent performance boost and cost reduction. With the reduced costs, we are able to compose our network with a hierarchical structure to process higher-resolution inputs.  We show that the proposed method with these integrated outperforms the previous state-of-the-art methods by large margins. Codes and pretrained weights are available at: \url{https://ku-cvlab.github.io/CATs-PlusPlus-Project-Page/}
\end{abstract}

}

\maketitle
\IEEEdisplaynontitleabstractindextext
\IEEEpeerreviewmaketitle



\IEEEraisesectionheading{\section{Introduction}\label{sec:introduction}}

\IEEEPARstart{E}{stablishing} dense correspondences across semantically similar images can facilitate many Computer Vision applications, including semantic segmentation~\cite{rubinstein2013unsupervised,taniai2016joint,min2021hypercorrelation}, object detection~\cite{lin2017feature}, and image editing~\cite{szeliski2006image,liu2010sift,liao2017visual,kim2019semantic,peebles2021gan}. Unlike classical dense correspondence problems such as stereo matching, geometric matching, or optical flow that consider visually similar images taken under the geometrically constrained settings~\cite{hosni2012fast,ilg2017flownet,sun2018pwc,hui2018liteflownet}, semantic correspondence poses additional challenges from large intra-class appearance and geometric variations~\cite{ham2016proposal,ham2017proposal,min2019spair} caused by the unconstrained settings of the given image pair.

Recent approaches~\cite{rocco2017convolutional,rocco2018end,rocco2018neighbourhood,melekhov2019dgc,min2019hyperpixel,min2020learning,liu2020semantic,truong2020glu,sarlin2020superglue,truong2020gocor,sun2021loftr,min2021convolutional} addressed these challenges by carefully designing deep Convolutional Neural Networks (CNNs)-based models analogously to the classical matching pipeline~\cite{scharstein2002taxonomy,philbin2007object}, namely feature extraction, cost aggregation, and flow estimation. Several works~\cite{kim2017fcss,detone2018superpoint, min2019hyperpixel, min2020learning,sarlin2020superglue,sun2021loftr} focused on the feature extraction stage, as it has been demonstrated that the more powerful feature representation the model learns, the more robust matching is obtained~\cite{kim2017fcss,detone2018superpoint,sun2021loftr}. However, solely relying on the matching similarity between features without any prior, \textit{e.g.,} spatial smoothness often suffers from the challenges due to ambiguities generated by repetitive patterns or background clutters~\cite{rocco2017convolutional,kim2017fcss,lee2019sfnet}. On the other hand, some methods~\cite{rocco2017convolutional,paul2018attentive,rocco2018end,kim2018recurrent,lee2019sfnet,truong2020glu} focused on the flow estimation stage either by designing additional CNN as an ad-hoc regressor that predicts the parameters of a single global transformation~\cite{rocco2017convolutional,rocco2018end}, finding confident matches from correlation maps~\cite{jeon2018parn,lee2019sfnet}, or directly feeding the correlation maps into the decoder to infer dense correspondences~\cite{truong2020glu}. However, these methods highly rely on the quality of the initial correlation maps~\cite{truong2020gocor}.

To address this, the latest methods~\cite{rocco2018neighbourhood,min2019hyperpixel,rocco2020efficient,jeon2020guided,liu2020semantic,li2020correspondence,min2021convolutional} have focused on the second stage, highlighting the importance of cost aggregation. Since the quality of correlation maps is of prime importance, they proposed to refine the matching scores by formulating the task as Optimal Transport problem~\cite{sarlin2020superglue,liu2020semantic}, re-weighting matching scores by Hough space voting for geometric consistency~\cite{min2019hyperpixel,min2020learning}, or utilizing high-dimensional 4D or 6D convolutions to find locally consistent matching points~\cite{rocco2018neighbourhood,rocco2020efficient,li2020correspondence,min2021convolutional,min2021efficient}. Although formulated variously, these methods either use hand-crafted techniques that are neither learnable nor robust to severe deformations, nor inherit the limitation of CNNs, which their limited receptive fields make it hard to discriminate incorrect matching points, and they lack an ability to adapt to the input contents due to the inherent 
characteristic of convolution, \textit{i.e.,} a fixed and shared kernel across all the input pixels.

In this work, we also focus on the cost aggregation stage and propose a novel cost aggregation network to tackle the aforementioned issues. Our network, called Cost Aggregation with Transformers (CATs), is based on transformer~\cite{vaswani2017attention,dosovitskiy2020image}, which is renowned for its global receptive field and ability to flexibly adapt to consider pairwise interactions among all the input tokens. By considering all the matching scores computed between features of input images globally, our aggregation network explores global consensus and thus refines the ambiguous or noisy matching scores effectively. Specifically, based on the observation that desired correspondence should be aligned at discontinuities with the appearance of images~\cite{zhang2014cross,kendall2017end}, we concatenate an appearance embedding to the correlation map, which helps to disambiguate the correlation map within the transformer aggregator. To benefit from hierarchical feature representations, following~\cite{lee2019sfnet,min2020learning,truong2020glu}, we use a stack of correlation maps constructed from multi-level features, and propose to effectively aggregate the scores across the multi-level correlation maps. Furthermore, we consider the bidirectional nature of a correlation map to leverage the correlation map from both source and target directions, obtaining reciprocal scores by swapping the pair of dimensions of the correlation map to allow global consensus. In addition to these, we provide residual connections around aggregation networks to ease the learning process.

\begin{figure}[t]
	\centering
	\renewcommand{\thesubfigure}{}
	\subfigure[(a)]
	{\includegraphics[width=0.495\linewidth]{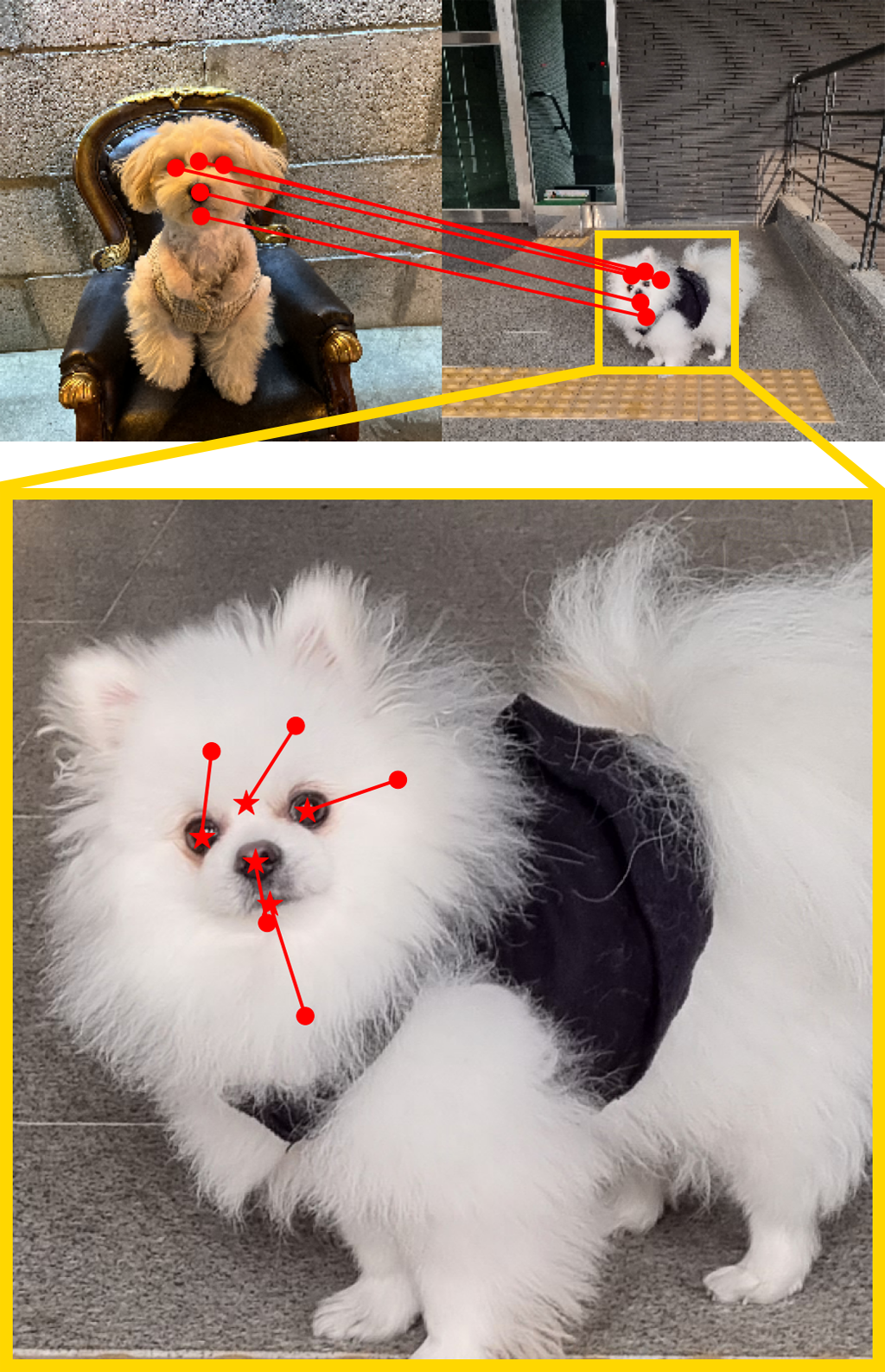}}\hfill
	\subfigure[(b)]
	{\includegraphics[width=0.495\linewidth]{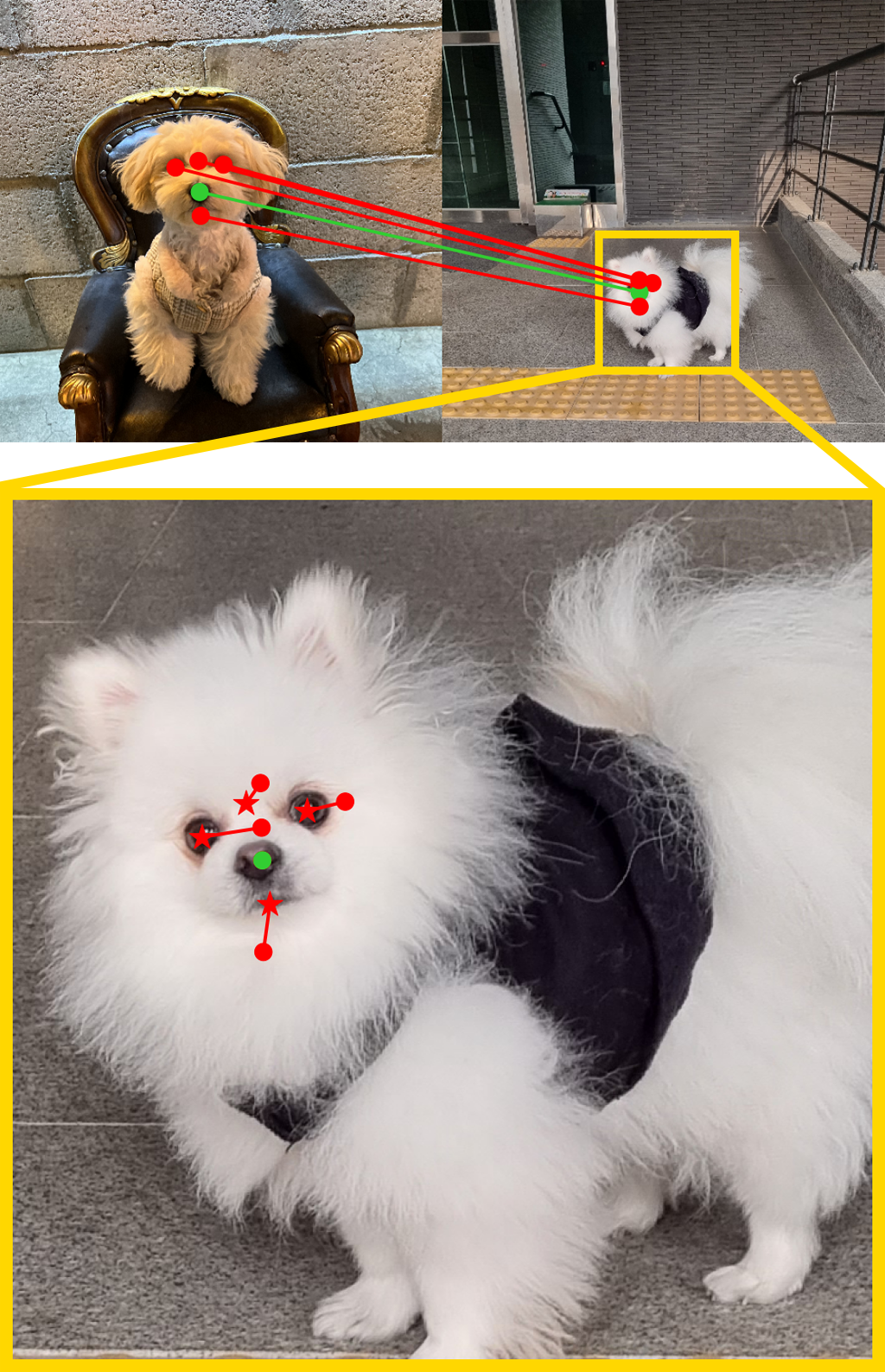}}\hfill\\
	\vspace{-10pt}
	\caption{\textbf{Qualitative comparison of (a) CATs and (b) CATs++.} We show that our proposed methods perform surprisingly well even for images in the wild. Note that green and red circles denote the correctly predicted points and incorrectly predicted points, respectively, and the red star denotes the ground-truth keypoint. CATs++ finds highly accurate and fine-grained matching points, even surpassing CATs, thanks to its hierarchical architecture as well as its enhanced cost aggregation components.   }\label{fig1}
	\vspace{-10pt}
\end{figure}  

Although competitive performance can be attained by CATs, it may face some limitations, \textit{i.e.,} high computational costs induced by the use of full attention adopted by standard transformers~\cite{vaswani2017attention}, which restrict its applicability only at limited resolution and result in rather limited performance. To address these, we propose CATs++ to not only alleviate the computational burden but also to enhance the cost aggregation for the improved performance, as exemplified in Fig.~\ref{fig1}. Specifically, we introduce early convolutions to reduce the costs by controlling the number of tokens and their dimensions and inject some convolutional inductive bias prior to cost aggregation by transformers, which also leads to apparent performance boost as this allows the model capable of aggregating both local and global interactions. Furthermore, we introduce a novel design for transformer, tailored for cost aggregation, which includes reformulation of standard Query-Key-Value (QKV) projection and Feed-Forward Network (FFN). This is achieved by including appearance affinity at the intermediate process of QKV projection to reduce computational burden and utilizing 4D convolutions rather than the linear projection in FFN to not only strengthen the power to consider locality but also to flexibly control the input dimensions. With the reduced costs from the reformulation,  we compose a hierarchical architecture, allowing the 
coarser levels to guide the cost aggregation at finer levels. Lastly, unlike CATs, we refrain from arbitrarily selecting different combinations of feature maps~\cite{min2019hyperpixel,li2020correspondence,cho2021semantic} for each dataset, which adds extra burden, but rather use all the feature maps as done in~\cite{min2021hypercorrelation} to exploit richer semantic representations.

With these combined, we demonstrate our methods on standard benchmarks~\cite{min2019spair,ham2016proposal,ham2017proposal} for semantic correspondence. Experimental results on various benchmarks prove the effectiveness of the proposed models over the latest methods for semantic correspondence, clearly outperforming and attaining state-of-the-art for all the benchmarks. We also provide an extensive ablation study to validate and analyze components. Finally, we show that the proposed method is robust to both domain and class shift, and also works surprisingly well for a few-shot segmentation task as one of the possible applications. 

This work is extended from the preliminary version of~\cite{cho2021semantic} by (\textbf{I}) leveraging both convolutions and transformers to take the best of two by strengthening the ability to consider locality interactions prior to global consideration of pairwise interactions by transformers and reduce the computational loads, (\textbf{II})  introducing a novel and efficient transformer architecture that can significantly boost the performance and reduce the computational costs, (\textbf{III}) designing a hierarchical architecture that aggregates the cost volume constructed at higher resolution, (\textbf{IV}) setting a new state-of-the-art performance for standard benchmarks~\cite{min2019spair,ham2016proposal,ham2017proposal} in semantic correspondence, making an 8.5$\%$p increase from the best-published results, and (\textbf{V}) providing additional experimental results, ablation studies and visualizations.


\section{Related Work}

\subsection{Semantic Correspondence}
Methods for semantic correspondence generally follow the classical matching pipeline~\cite{scharstein2002taxonomy,philbin2007object}, including feature extraction, cost aggregation, and flow estimation. Most early efforts~\cite{dalal2005histograms,liu2010sift,ham2016proposal} leveraged the hand-crafted features which are inherently limited in capturing high-level semantics. Though using deep CNN-based features~\cite{choy2016universal,kim2017fcss,rocco2017convolutional,rocco2018end,kim2018recurrent,paul2018attentive,lee2019sfnet} has become increasingly popular thanks to their invariance to deformations, without a means to refine the matching scores independently computed between the features, the performance would be rather limited. Recently, MMNet~\cite{zhao2021multi} attempts to use transformers~\cite{vaswani2017attention} to refine the features, focusing on the feature extraction stage, with slight changes made to self-attention computation to compute local self-attention over feature maps to aggregate the local feature values, while we take a different approach, aggregating the cost volume and leveraging both convolutions and transformers to not only inject some convolutional inductive bias and reduce computation cost but also to exploit the merits of both, {\it i.e.,} locality bias of convolution, adaptability and global receptive fields of transformers.

To alleviate the limited performance by the use of matching scores without aggregation, several methods focused on the flow estimation stage. Rocco et al.~\cite{rocco2017convolutional,rocco2018end} proposed an end-to-end network to predict global transformation parameters from the matching scores, and their success inspired many variants~\cite{paul2018attentive,kim2018recurrent,kim2019semantic}. Among the subsequent works, RTNs~\cite{kim2018recurrent} attempted to obtain semantic correspondences through an iterative process of estimating spatial transformations. DGC-Net~\cite{melekhov2019dgc}, Semantic-GLU-Net~\cite{truong2020glu}, and DMP~\cite{Hong_2021_ICCV} utilize a CNN-based decoder to directly find correspondence fields. Recently, PDC-Net~\cite{truong2021learning} proposed a flexible probabilistic model that jointly learns the flow estimation and its uncertainty. Arguably, directly regressing correspondences from the initial matching scores highly relies on the quality of them, which motivates aggregation of matching scores for more reliable correspondences.

Recent numerous methods~\cite{rocco2018neighbourhood,min2019hyperpixel,min2020learning,liu2020semantic,sarlin2020superglue,sun2021loftr,min2021convolutional} thus have focused on the cost aggregation stage to refine the initial matching scores. Among hand-crafted methods, SCOT~\cite{liu2020semantic} formulates semantic correspondence as an Optimal Transport problem and attempts to solve two issues, namely many to one matching and background matching. HPF~\cite{min2019hyperpixel} first computes appearance matching confidence using hyperpixel features and then uses Regularized Hough Matching (RHM) algorithm  for cost aggregation to enforce geometric consistency. DHPF~\cite{min2020learning}, which replaces the feature selection algorithm of HPF~\cite{min2019hyperpixel} with trainable networks, also uses RHM. However, these hand-crafted techniques for refining the matching scores are neither learnable nor robust to severe deformations. 

As learning-based approaches, NC-Net~\cite{rocco2018neighbourhood} utilizes 4D convolution to achieve local neighborhood consensus by finding locally consistent matches, and its variants~\cite{rocco2020efficient,li2020correspondence} proposed more cost-efficient methods. PMNC~\cite{lee2021patchmatch} utilizes a classical algorithm, PatchMatch~\cite{barnes2009patchmatch}, and 4D convolutions to tackle the semantic correspondence task, and proposes a PahchMatch-based optimization strategy to iteratively refine the correspondence field. GOCor~\cite{truong2020gocor} proposed an aggregation module that directly improves the correlation maps. GSF~\cite{jeon2020guided} formulated a pruning module to suppress false positives of correspondences in order to refine the initial correlation maps. CHM~\cite{min2021convolutional} goes one step further, proposing a learnable geometric matching algorithm that utilizes 6D convolution. CHMNet~\cite{min2021efficient} extends CHM~\cite{min2021convolutional} by introducing an efficient kernel decomposition with center-pivot neighbors. Although outstanding performance with the help of additional data augmentation~\cite{cho2021semantic} and multi-level features~\cite{min2019hyperpixel,cho2021semantic,min2020learning} is achieved, CHMNet~\cite{min2021efficient} yet suffers from limitations of CNN-based architectures for cost aggregation.

\subsection{Visual Correspondence Applications}
Establishing visual correspondences has been a cornerstone of many Computer Vision applications. To name a few, object tracking~\cite{Gao2022AiATrackAI,yan2022towards}, video object segmentation (VOS)~\cite{wang2019learning,lai2019self,jabri2020space}, few-shot segmentation (FSS)~\cite{hong2022cost,min2021hypercorrelation}, visual localization~\cite{li2020dual,rocco2020efficient}, structure-from-motion (SfM)~\cite{schonberger2016structure}, 
and image retrieval~\cite{tan2021instance,lee2022cvnet} are the tasks that establishing accurate correspondence fields remain as one of the milestones for their goals. More specifically, object tracking~\cite{yilmaz2006object} and video object segmentation~\cite{perazzi2016benchmark} need space-time correspondences between objects of interests for detection or segmentation across different time steps, and FSS~\cite{dong2018few} requires segmenting objects of unseen classes by finding correspondences given a few annotated support images and a query image. SfM~\cite{schonberger2016structure} and visual localization~\cite{lee2022cvnet} are the 3D computer vision tasks to estimate three-dimensional structures 
by finding correspondences across multiple two-dimensional image sequences. In this paper, we also show how the proposed method is applicable to FSS and visual localization tasks, and evaluate its effectiveness.

\subsection{Transformers in Vision}
Transformers~\cite{vaswani2017attention}, the \emph{de facto} standard for Natural Language Processing (NLP) tasks, has recently imposed significant impact on various tasks in Computer Vision fields such as image classification~\cite{dosovitskiy2020image,touvron2020deit}, object detection~\cite{carion2020end,zhu2020deformable}, tracking and matching~\cite{sun2020transtrack,sun2021loftr}. ViT~\cite{dosovitskiy2020image}, the first work to propose an end-to-end transformer-based architecture for the image classification task, successfully extended the receptive field, owing to its self-attention nature that can capture the global relationships between features. For those works addressing visual correspondence, LoFTR~\cite{sun2021loftr} uses a cross and self-attention module to refine the feature maps conditioned on both input images, and formulate the hand-crafted aggregation layer with dual-softmax~\cite{rocco2018neighbourhood,tyszkiewicz2020disk}, and Optimal Transport~\cite{sarlin2020superglue} to infer correspondences. In another work, COTR~\cite{jiang2021cotr} takes coordinates as input and addresses dense correspondence tasks without the use of a correlation map. Unlike these, for the first time, we propose a transformer-based cost aggregation module. 

\begin{figure*}
    \centering
    \includegraphics[width=0.9\linewidth]{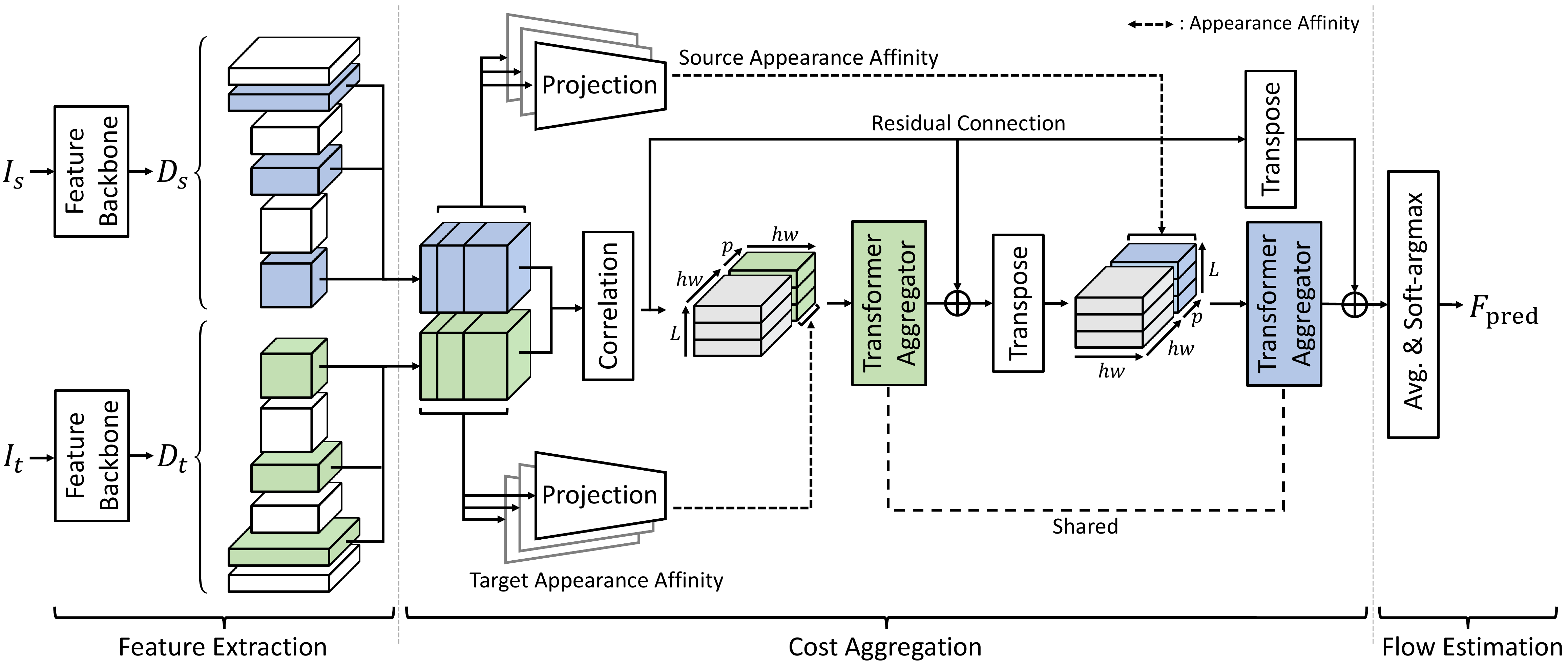}\hfill\\
    \caption{\textbf{Overall network architecture of CATs.} Analogously to the classical matching pipeline, our networks consist of feature extraction, cost aggregation, and flow estimation modules. We first extract multi-level dense features and construct a stack of correlation maps. We then concatenate with embedded features and feed into the transformer-based cost aggregator to obtain a refined correlation map. The flow is then inferred from the refined correlation map.}
    \label{fig:catsarch}\vspace{-10pt}
\end{figure*}


\section{Methodology}\label{sec:CHM}

\subsection{Motivation and Overview}

Let us denote a pair of images as source $I_{s}$ and  target $I_{t}$, which represent semantically similar images, and features extracted from $I_{s}$ and $I_{t}$ as $D_{s}$ and $D_{t}$, respectively. Here, our goal is to establish a dense correspondence field $F(i)$ between two images that are defined for each pixel $i$, which warps $I_{t}$ towards $I_{s}$.

Estimating the correspondence with sole reliance on matching similarities between $D_{s}$ and $D_{t}$ is often challenged by ambiguous matches due to the repetitive patterns or background clutters~\cite{rocco2017convolutional,kim2017fcss,lee2019sfnet,truong2020glu,Hong_2021_ICCV}. To address this, numerous methods proposed cost aggregation techniques that try to refine the initial matching similarities either by formulating the task as Optimal Transport problem~\cite{sarlin2020superglue,liu2020semantic}, using regularized Hough Matching to re-weight the costs~\cite{min2019hyperpixel,min2020learning} or adopting 4D or 6D convolutions~\cite{rocco2018neighbourhood,li2020correspondence,rocco2020efficient,min2021convolutional,min2021efficient}. However, these methods either use hand-crafted techniques that are weak to severe deformations or only consider long-range pairwise interactions {implicitly} by stacking the series of convolution blocks. 

To overcome these, we present transformer-based cost aggregation networks that effectively integrate information present in all matching costs as a whole by considering pairwise interactions  {explicitly} with the help from global receptive fields of attention mechanism, as illustrated in Fig.~\ref{fig:catsarch}. In the following, we explain feature extraction and cost computation and then explain the proposed transformer aggregator. Finally, we introduce additional techniques to further boost the performance to complete the design of cost aggregation networks with transformers.

\subsection{Feature Extraction and Cost Computation}
\begin{figure*}[t]
\centering
\renewcommand{\thesubfigure}{}
\subfigure[(a) Source]
{\includegraphics[width=0.123\textwidth]{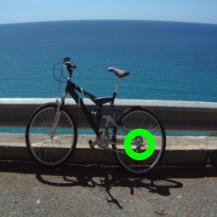}}\hfill
\subfigure[(b) Target]
{\includegraphics[width=0.123\textwidth]{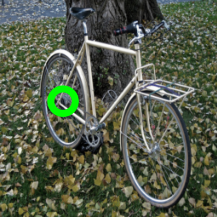}}\hfill
\subfigure[(c) Raw Corr. \#2 ]
{\includegraphics[width=0.123\textwidth]{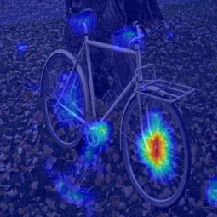}}\hfill
\subfigure[(d) Raw Corr. \#4]
{\includegraphics[width=0.123\textwidth]{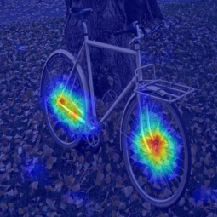}}\hfill
\subfigure[(e) Raw Corr. \#5]
{\includegraphics[width=0.123\textwidth]{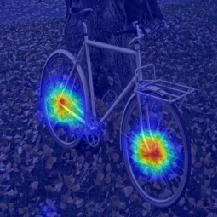}}\hfill
\subfigure[(f) Raw Corr. \#7]
{\includegraphics[width=0.123\textwidth]{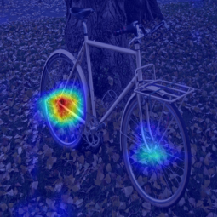}}\hfill
\subfigure[(g) HPF~\cite{min2019hyperpixel}]
{\includegraphics[width=0.123\textwidth]{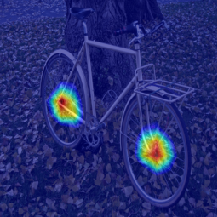}}\hfill
\subfigure[(h) CATs]
{\includegraphics[width=0.123\textwidth]{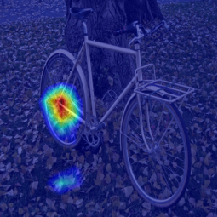}}\hfill\\
\vspace{-10pt}
\caption{\textbf{Visualization of multi-level aggregation:} (a) source, (b) target images, (c)-(f) raw correlation maps, respectively, and final correlation maps by (e) HPF~\cite{min2019hyperpixel} and (f) CATs. Note that HPF and CATs utilize the same feature maps. Compared to HPF, CATs successfully embrace richer semantics in different levels of feature maps.   }
\label{fig:multi-cost}\vspace{-10pt}
\end{figure*}

To extract dense feature maps from images, as shown in~\cite{min2019hyperpixel,melekhov2019dgc,min2020learning,truong2020glu,liu2020semantic}, we leverage multi-level features to capture hierarchical semantic feature representations. To this end, we use multi-level features from different levels of convolutional layers to construct a stack of correlation maps.
Specifically, we use CNNs\footnote[1]{Note that we could also use transformer-based feature backbone networks, \textit{e.g.,} ViT~\cite{dosovitskiy2020image}, which we explore the influence of different feature backbone networks in Section 5.4.} that produce a sequence of $L$ feature maps, and $D^{l}$ represents a feature map at the $l$-th level. As done in~\cite{min2019hyperpixel,liu2020semantic}, we use a different combination of multi-level features depending on the dataset trained on, e.g., PF-PASCAL~\cite{ham2017proposal} or SPair-71k~\cite{min2019spair}. Given a sequence of feature maps, we resize all the selected feature maps to $\mathbb{R}^{h\times w \times c}$, with height $h$, width $w$, and $c$ channels. Note that we down-sample the features to fixed $h$ and $w$ for efficiency. The resized features first undergo $l$-2 normalization and a correlation map is computed using the inner product between them followed by an activation function such that: 
\begin{equation}
    \mathcal{C}^l(i,j)=\mathrm{ReLU}\left(
    \frac{D^l_{s}(i)\cdot {D}^l_{t}(j)}{\|D^l_{s}(i)\|\|{D}^l_{t}(j)\|}\right),
\end{equation}
where $i$ and $j$ denote 2D spatial positions of feature maps, respectively,  and $\mathrm{ReLU}(\cdot)$ denotes ReLU activation function~\cite{agarap2018deep}. In this way, all pairwise feature matches are computed and stored. We subsequently concatenate the computed correlation maps to obtain a stacked correlation map $\mathcal{C} \in \mathbb{R}^{hw \times hw \times L}$. However, raw matching scores contain numerous ambiguous matching points as exemplified in Fig.~\ref{fig:multi-cost}, which results in inaccurate correspondences. To remedy this, we propose cost aggregation networks in the following that aim to refine the ambiguous or noisy matching scores.

\subsection{Preliminary: Transformers}
Before moving on to the proposed transformer aggregator, we start by explaining Vision Transformer (ViT) encoder~\cite{vaswani2017attention}. Note that the proposed method does not utilize a decoder, which we omit its explanation in this section. Renowned for its global receptive fields, one of the key elements of transformer~\cite{vaswani2017attention} is the self-attention mechanism, which enables finding the correlated input tokens by first feeding into scaled dot product attention function (self-attention layer~\cite{vaswani2017attention}), normalizing with Layer Normalization (LN)~\cite{ba2016layer}, and passing the normalized values to an MLP, which is also known as Feed-Forward Network (FFN). Formulating the overall process of vision transformer~\cite{dosovitskiy2020image}, $y = \mathcal{T}({I})$, where $y$ serves as the image representation obtained by prepending a learnable embedding prior to feeding into the encoder, it first undergoes Query-Key-Value projection as:
\begin{equation}
\begin{split}
        &Q = \mathcal{P}_Q(\mathrm{LN}(I' + E_{\mathrm{pos}})),\\
        &K = \mathcal{P}_K(\mathrm{LN}(I' + E_{\mathrm{pos}})),\\
        &V = \mathcal{P}_V(\mathrm{LN}(I' + E_{\mathrm{pos}})),
\end{split}
\end{equation}
where $\mathcal{P}_Q$, $\mathcal{P}_K$ and $\mathcal{P}_V$ denote query, key and value linear projection, respectively; $I'$ denotes token embeddings; $E_{\mathrm{pos}}$ denotes positional embedding. Then the obtained $Q$, $K$, and $V$ undergo a self-attention layer:
\begin{equation}
    \hat{z} = \mathrm{SA}(Q, K, V) = \mathrm{softmax}(QK^T)V,
    \label{selfatt}
\end{equation}
where $\mathrm{SA}(\cdot)$ denotes Self-Attention. Subsequently, the output undergoes LN and residual connection followed by FFN that consists of two linear transformations with a GELU~\cite{hendrycks2016gaussian} activation in between, as shown in Fig.~\ref{efficient}, which is formulated as:
\begin{equation}
\begin{split}
    &{z} = \hat{z} + z_\mathrm{pos},\\
    &x = \mathrm{LN}(z),\\
    &\hat{x} = \mathcal{P}_{\mathrm{FFN}}^2(\mathrm{GELU}(\mathcal{P}_{\mathrm{FFN}}^1(x))),\\
    &y = \hat{x} + {z},
    \label{ffn}
\end{split}
\end{equation}
where $z_\mathrm{pos} = I' + E_{\mathrm{pos}}$, and $\mathcal{P}_{\mathrm{FFN}}^1$ and $\mathcal{P}_{\mathrm{FFN}}^2$ are the first and the second position-wise linear transformations within the feed-forward network, respectively.

\begin{figure*}
    \centering
    \includegraphics[width=1.0\linewidth]{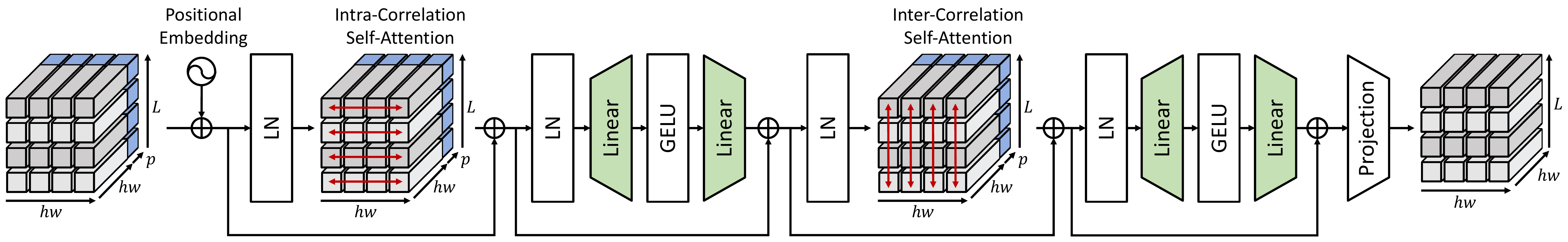}\hfill\\
    \caption{\textbf{Illustration of Transformer aggregator.} Given correlation maps $\mathcal{C}$ with projected features, transformer aggregator consisting of intra- and inter-correlation self-attention with LN and MLP refines the inputs not only across spatial domains but across levels.}
    \label{fig:aggregator}\vspace{-10pt}
\end{figure*}

\subsection{Transformer Aggregator}
\begin{figure*}[t]
\centering
\renewcommand{\thesubfigure}{}
\subfigure[]
{\includegraphics[width=0.123\textwidth]{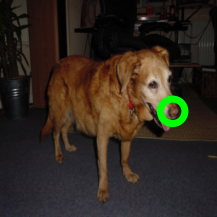}}\hfill
\subfigure[]
{\includegraphics[width=0.123\textwidth]{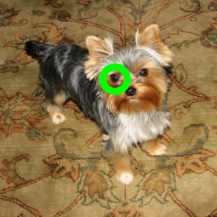}}\hfill
\subfigure[]
{\includegraphics[width=0.123\textwidth]{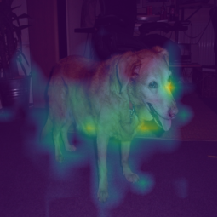}}\hfill
\subfigure[]
{\includegraphics[width=0.123\textwidth]{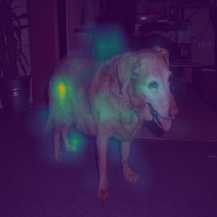}}\hfill
\subfigure[]
{\includegraphics[width=0.123\textwidth]{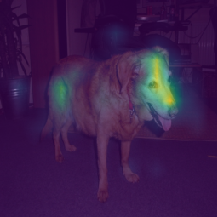}}\hfill
\subfigure[]
{\includegraphics[width=0.123\textwidth]{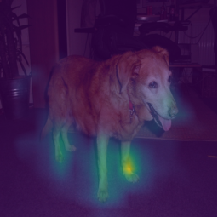}}\hfill
\subfigure[]
{\includegraphics[width=0.123\textwidth]{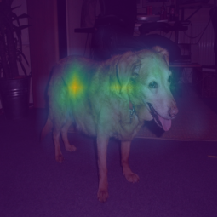}}\hfill
\subfigure[]
{\includegraphics[width=0.123\textwidth]{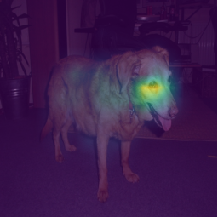}}\hfill\\
\vspace{-20.5pt}
\subfigure[(a) Source]
{\includegraphics[width=0.123\textwidth]{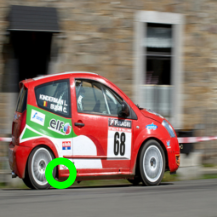}}\hfill
\subfigure[(b) Target]
{\includegraphics[width=0.123\textwidth]{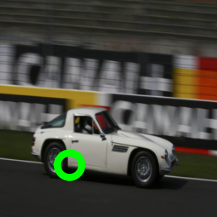}}\hfill
\subfigure[(c) Source $l$ = 1]
{\includegraphics[width=0.123\textwidth]{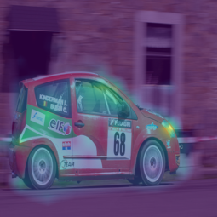}}\hfill
\subfigure[(d) Source $l$ = 2]
{\includegraphics[width=0.123\textwidth]{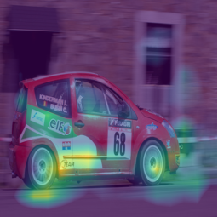}}\hfill
\subfigure[(e) Source $l$ = 3]
{\includegraphics[width=0.123\textwidth]{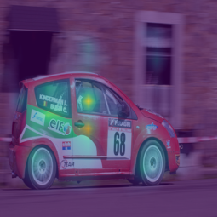}}\hfill
\subfigure[(f) Source $l$ = 4]
{\includegraphics[width=0.123\textwidth]{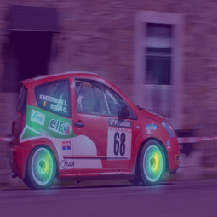}}\hfill
\subfigure[(g) Source $l$ = 5]
{\includegraphics[width=0.123\textwidth]{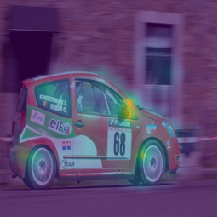}}\hfill
\subfigure[(h) Source $l$ = 6]
{\includegraphics[width=0.123\textwidth]{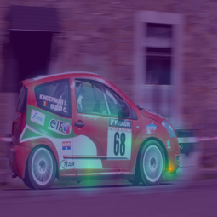}}\hfill\\
\vspace{-10pt}
\caption{\textbf{Visualization of self-attention:} Each attention map attends to different aspects, and CATs aggregates the costs, leveraging rich semantic representations. }
\label{fig:self-vis}\vspace{-10pt}
\end{figure*}

Several works~\cite{dosovitskiy2020image,carion2020end,zhu2020deformable,sun2021loftr} have shown that given images or features as input, transformers~\cite{vaswani2017attention} integrate the global context information by learning to find the attention scores for all pairs of tokens. In this paper, we take a different approach, leveraging the transformers to integrate the pairwise relationships across matching scores to not only discover global consensus  but also to carefully consider the pairwise interactions among tokens for an effective cost aggregation in a global manner. Unlike ViT~\cite{dosovitskiy2020image}, we do not use a class token and obtain a refined cost $\hat{\mathcal{C}}$ as output by feeding the stacked raw cost $\mathcal{C} \in \mathbb{R}^{hw \times hw \times L}$ to the transformer $\mathcal{T}$, consisting of QKV projections, SA, LN, and FFN:
\begin{equation}
    \hat{\mathcal{C}} = \mathcal{T}(\mathcal{C}).
\end{equation}
The standard transformers receive input as a 1D sequence of token embeddings. To this end, we also treat each spatial location at either source or target as 1D token. We visualize the refined correlation map with self-attention in Fig.~\ref{fig:self-vis}, where the ambiguities are significantly resolved. In the following, we introduce appearance affinity modeling to disambiguate the noisy correlation maps and multi-level aggregation to effectively aggregate across spatial- and level-dimension.

\subsubsection{Appearance Affinity Modelling}
When only matching costs are considered for aggregation, self-attention layer processes the correlation map disregarding the noise involved in the correlation map, which may lead to inaccurate correspondences. Rather than solely relying on a raw correlation map, we additionally provide an appearance embedding from input features to disambiguate the correlation map. The intuition behind is that visually similar points in an image, e.g., color or feature, have similar correspondences, as proven in stereo matching literature, e.g., Cost Volume Filtering (CVF)~\cite{hosni2012fast,sun2018pwc}. 

To provide appearance affinity, we propose to concatenate embedded features projected from input features with the correlation map. We first feed the features $D^l$ into the linear projection networks, and then concatenate the output along the corresponding dimension, so that the correlation map is augmented such that $[\mathcal{C}^l ,\mathcal{P}^l(D^l)] \in \mathbb{R}^{hw \times (hw + p) }$, where $[\,\cdot\,]$ denotes concatenation, $\mathcal{P}$ denotes linear projection networks, and $p$ is channel dimension of the embedded feature. 
The transformer can be then formulated as
\begin{equation}
    \hat{\mathcal{C}} = \mathcal{T}([\mathcal{C}^l, \mathcal{P}^l(D^l)]^L_{l=1}).
\end{equation}

Specifically, within the transformer $\mathcal{T}$, this augmented correlation map then undergoes QKV projection: 
\begin{equation}
\begin{split}
        &Q = \mathcal{P}_Q(\mathrm{LN}([\mathcal{C}^l ,\mathcal{P}^l(D^l)]^L_{l=1} + E_\mathrm{pos})),\\
        &K = \mathcal{P}_K(\mathrm{LN}([\mathcal{C}^l,\mathcal{P}^l(D^l)]^L_{l=1} + E_\mathrm{pos})),\\
        &V = \mathcal{P}_V(\mathrm{LN}([\mathcal{C}^l,\mathcal{P}^l(D^l)]^L_{l=1} + E_\mathrm{pos})),
\end{split}
\end{equation}
where for $E_\mathrm{pos}$, we let the networks be aware of the positional information by providing a learnable embedding~\cite{dosovitskiy2020image} rather than a fixed~\cite{vaswani2017attention} as shown in Fig.~\ref{fig:aggregator}.  Note that $\mathcal{P}_Q, \mathcal{P}_K$ and $\mathcal{P}_V$ process the stacked correlation map $\mathcal{C}$, while $\mathcal{P}$ processes at each $l$-th level of $\mathcal{C}$.
After the QKV projections, following the process of a standard transformer, we feed the output into the self-attention layer, which is formulated in Eq.~\ref{selfatt}, to aggregate the matching costs concatenated with appearance embedding to reason about their relationships, and then they undergo FFN, which is formulated in Eq.~\ref{ffn}. Subsequently, the aggregated augmented correlation map is passed to the linear projection networks, to retain the size of original correlation $\mathcal{C}$, which we obtain $\hat{\mathcal{C}}$ as illustrated in Fig.~\ref{fig:aggregator}.

\subsubsection{Multi-level Aggregation}
For each $l$-th level of stacked correlation maps $\mathcal{C} = [\mathcal{C}^l]^L_{l=1}$, level-wise matching scores are available, and we aim to employ all of them present across different levels. In order to employ both hierarchical and spatial semantic representations, we introduce a multi-level aggregation technique, which aggregates the stack of augmented correlation maps in a spatial- and level-wise manner. In this way, the aggregation networks can consider interactions among matching scores at a particular level, as well as across levels.  

Specifically, as shown in Fig.~\ref{fig:aggregator}, a stack of $L$ augmented correlation maps, $[\mathcal{C}^l, \mathcal{P}^l(D^l)]^L_{l=1} \in \mathbb{R}^{hw \times (hw + p) \times L}$, undergo the transformer aggregator. For each $l$-th augmented correlation map, we aggregate with self-attention layer across all the points in the augmented correlation map, and we refer to this as {\it intra}-correlation self-attention. This is shown as a red arrow, which is a level-wise aggregation. In addition, subsequent to this, the correlation map undergoes {\it inter}-correlation self-attention across multi-level dimensions, which is a spatial-wise aggregation. Contrary to HPF~\cite{min2019hyperpixel} that concatenates all the multi-level features to compute a stack of correlation maps and aggregate using RHM, which only limitedly consider the multi-level similarities, within the inter-correlation self-attention layer of the proposed model, the similar matching scores are explicitly explored across multi-level dimensions. In this way, we can embrace richer semantics in different levels of feature maps, as shown in Fig.~\ref{fig:multi-cost}. 

\subsection{Cost Aggregation with Transformers}
By leveraging the transformer aggregator, we present a cost aggregation framework with the following additional techniques to improve the performance, and by combining them, we complete the architecture of CATs.
\subsubsection{Swapping Self-Attention} 
In order to impose mutually consistent matching scores, we argue that reciprocal scores should be used to infer confident correspondences. As the correlation map contains bidirectional matching scores, from both target and source perspectives, we can leverage matching similarities from both directions in order to obtain more reciprocal scores as done similarly in other works~\cite{rocco2018neighbourhood,lee2019sfnet}.

As shown in Fig.~\ref{fig:catsarch}, we first feed the augmented correlation map to the aforementioned transformer aggregator. Then we transpose the output, swapping the pair of dimensions in order to concatenate with the embedded feature from the other image and feed it into the subsequent aggregator. Note that we share the parameters of the transformer aggregators to obtain reciprocal scores. Formally, we define the whole process as follows:
\begin{equation}
\begin{split}
   &\mathcal{S} = \mathcal{T}([\mathcal{C}^l, \mathcal{P}^l(D^l_t)]^L_{l=1}),\\
   &\hat{\mathcal{C}} = \mathcal{T}([(\mathcal{S}^l)^\mathrm{T} , \mathcal{P}^l(D^l_s)]^L_{l=1}),    
   \label{eq7}
\end{split}
\end{equation}
where $\mathcal{C}^\mathrm{T}{(i,j)}=\mathcal{C}{(j,i)} $ denotes swapping the pair of dimensions corresponding to the source and target images; $\mathcal{S}$ denotes the intermediate correlation map before swapping the axis. Note that NC-Net~\cite{rocco2018neighbourhood} proposed a similar procedure, but instead of processing serially, they separately process the correlation map and its transposed version and add the outputs, which this procedure is designed to produce a correlation map invariant to the particular order of the input images. Unlike this, we process the correlation map serially, first aggregating the one pair of dimensions and then further aggregating the other pair. In this way, the subsequent attention layer is given cost maps with more consistent matching scores as an input, allowing further reduction of inconsistent matching scores. We justify our design choice in the ablation study.  

\begin{figure*}
    \centering
    \includegraphics[width=0.9\linewidth]{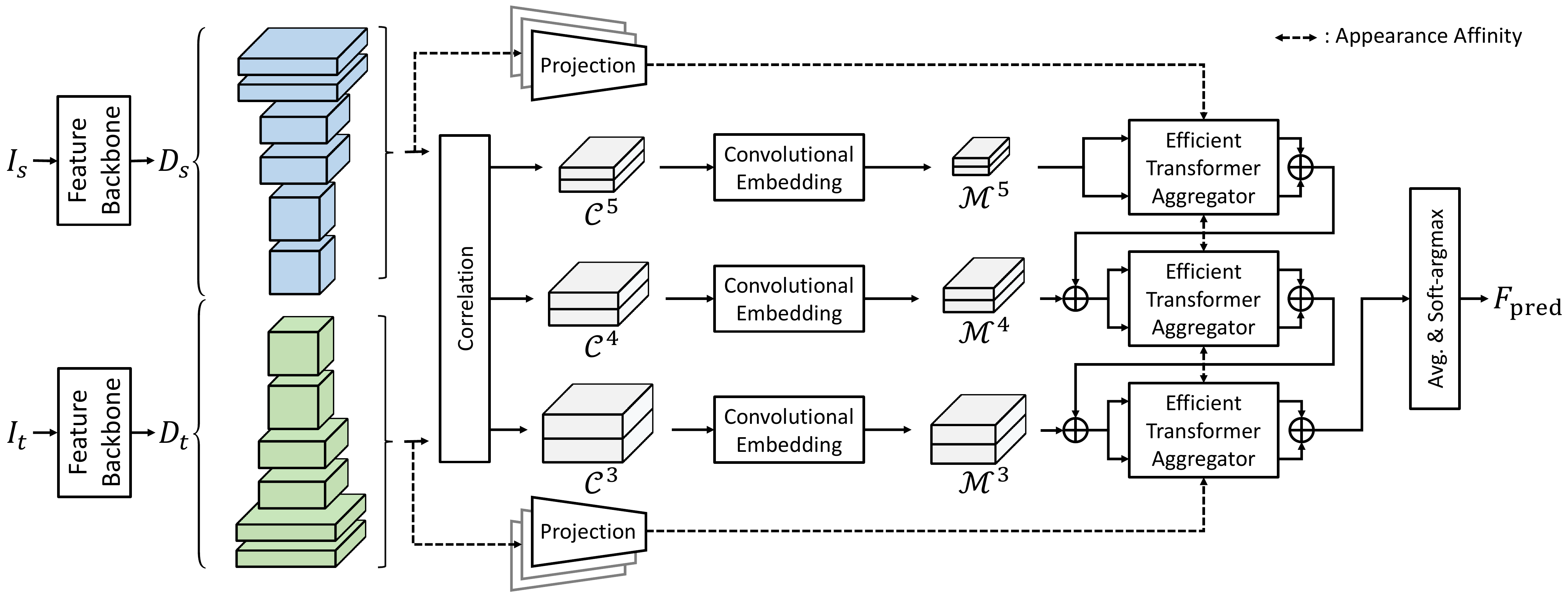}\hfill\\
    \caption{\textbf{Overall network architecture of CATs++.} The networks of CATs++ differ from CATs in that all the feature maps are used to compute cost volumes rather than selecting different combinations for different datasets, and early convolution along with the efficient transformer aggregator helps to reduce computational costs and improve performance. Note that we colored the extracted feature maps to indicate that we use all the intermediate feature maps and highlight the difference to Fig.~\ref{fig:catsarch}.}
    \label{fig:CATs++}\vspace{-10pt}
\end{figure*}

\subsubsection{Residual Connection} At the initial phase when the correlation map is fed into the transformers, noisy score maps are inferred due to randomly initialized parameters, which could complicate the learning process. To stabilize the learning process and provide a better initialization for the matching, we employ the residual connection. Specifically, we enforce the cost aggregation networks to estimate the residual correlation by adding residual connection around aggregation networks. Now Eq.~\ref{eq7} changes to:
\begin{equation}\label{eq2}
\begin{split}
   &\mathcal{S} = \mathcal{T}([\mathcal{C}^l, \mathcal{P}^l(D^l_t)]^L_{l=1}) + \mathcal{C},\\
   &\hat{\mathcal{C}} = \mathcal{T}([(\mathcal{S}^l)^\mathrm{T} , \mathcal{P}^l(D^l_s)]^L_{l=1})+ \mathcal{C}^T.  
\end{split}
\end{equation}

\subsection{Training Objective}
As in~\cite{min2019hyperpixel,min2020learning,min2021convolutional}, we assume that the ground-truth keypoints are given for each pair of images. We first average the stack of refined correlation maps $\hat{\mathcal{C}} \in \mathbb{R}^{hw \times hw \times L} $ along level dimension and then transform it into a dense flow field $F_\mathrm{pred}$ using soft-argmax operator~\cite{lee2019sfnet}. Subsequently, we compare the predicted dense flow field with the ground-truth flow field $F_\mathrm{GT}$ obtained by following the protocol of~\cite{min2019hyperpixel} using input keypoints. For the training objective, we utilize Average End-Point Error (AEPE)~\cite{melekhov2019dgc}, computed by averaging the Euclidean distance between the ground-truth and estimated flow. We thus formulate the objective function as $\mathcal{L}=\|F_\mathrm{GT}-F_\mathrm{pred}\|_{2}$.

\begin{figure*}[t]
	\centering
	\renewcommand{\thesubfigure}{}
	\subfigure[(a)]
	{\includegraphics[width=0.366\linewidth]{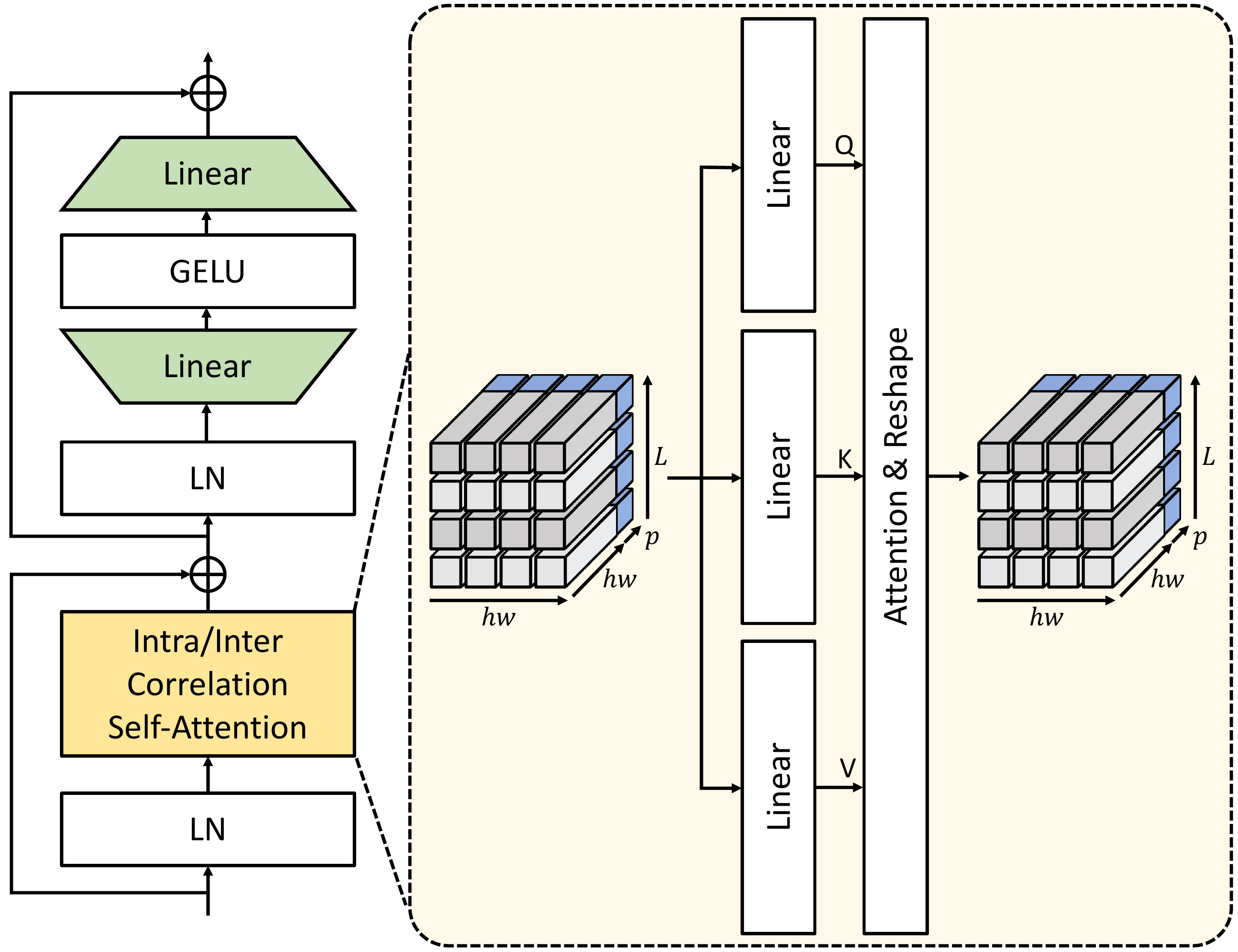}}\hfill
	\subfigure[(b)]
	{\includegraphics[width=0.618\linewidth]{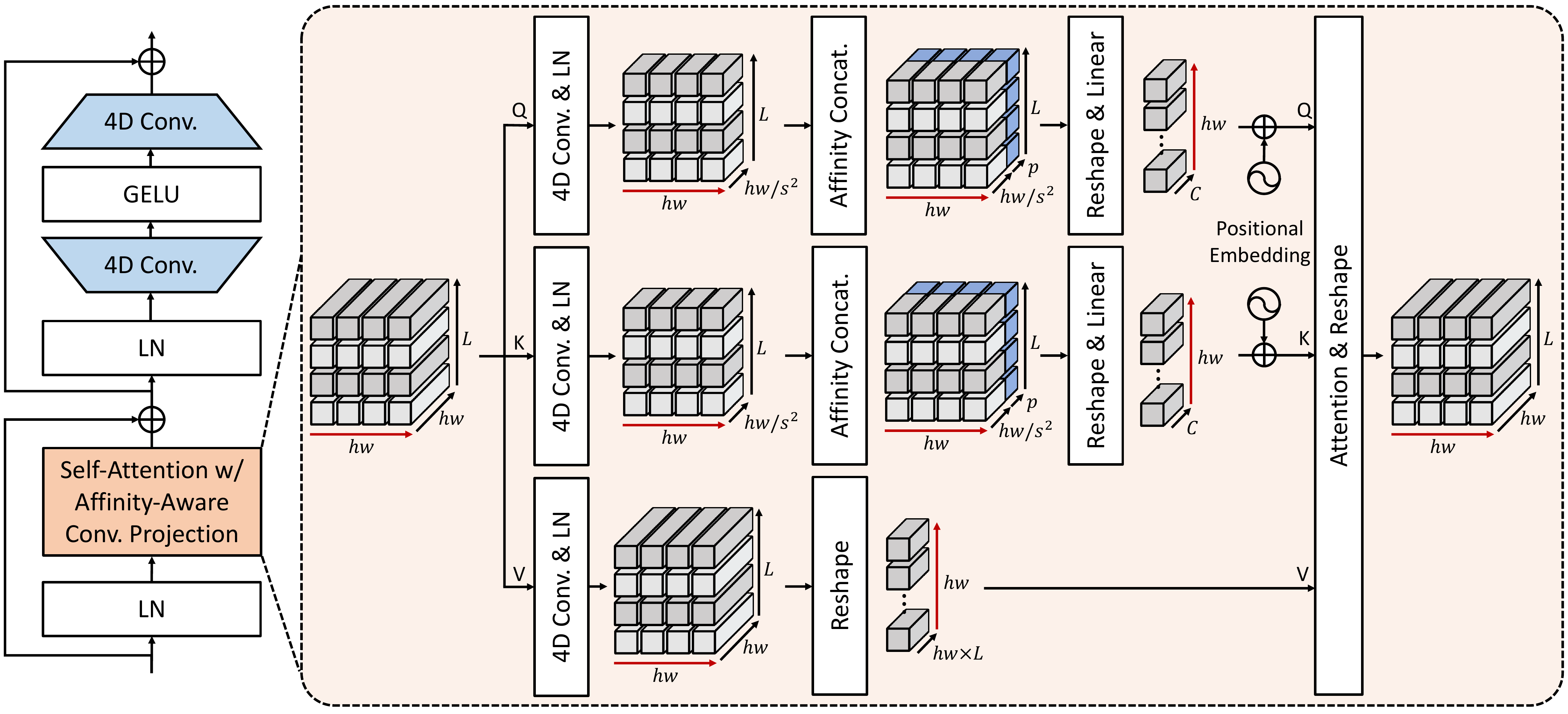}}\hfill\\
	\vspace{-10pt}
	\caption{\textbf{Intuition of the proposed components:} (a) standard transformer-based aggregator (in CATs), (b) efficient transformer-based aggregator (in CATs++). CATs++ replaces the query, key and value projections and feed-forward networks of standard transformer aggregator. With this novel design, we achieve significant performance boost as well as meaningful costs reductions.    }\label{efficient}\vspace{-10pt}
\end{figure*}  

\section{Boosting Cost Aggregation with Convolutions and Transformers}
\subsection{Motivation and Overview}
Despite its deliberate design and effectiveness, CATs may struggle with high computational costs induced by the use of standard transformers ~\cite{vaswani2017attention}, and this could restrict the cost aggregation to be performed only at limited resolutions. More specifically, as stated in some works~\cite{dosovitskiy2020image,vaswani2017attention,marin2021token} that computational complexity for self-attention layer is quadratic to token length, and those of QKV projection and FFN are quadratic to feature dimension, processing high-dimensional correlation maps with standard transformers~\cite{vaswani2017attention} inevitably face large computational burden. 

\begin{figure}
	\centering
	\renewcommand{\thesubfigure}{}
	\subfigure[(a) Patch embedding]
	{\includegraphics[width=0.47\linewidth]{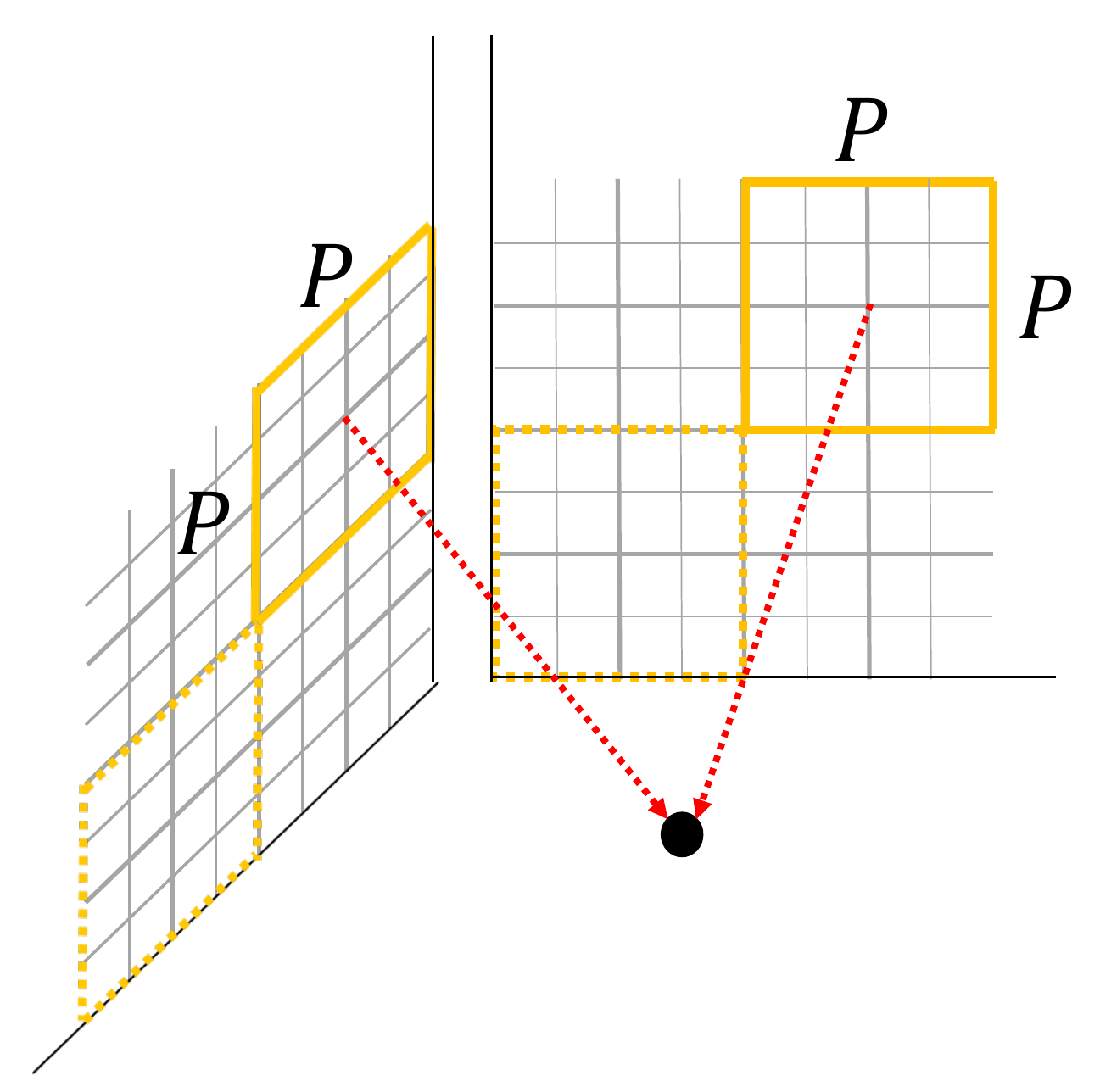}}\hfill
	\subfigure[(b) Convolutional embedding ]
	{\includegraphics[width=0.47\linewidth]{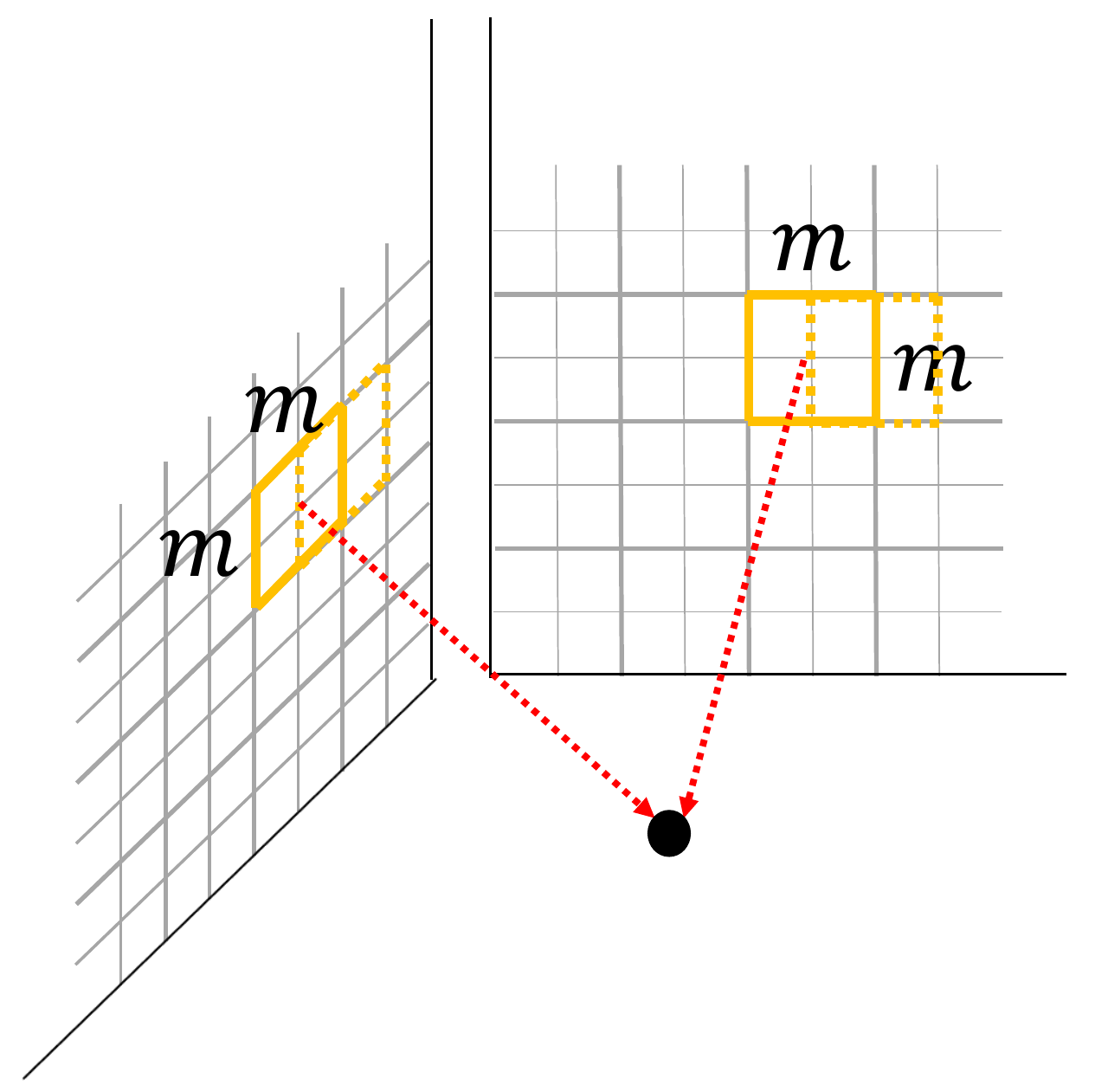}}\hfill\\\vspace{-10pt}
    \caption{\textbf{Comparison of different embedding strategy.} Unlike patch embedding~\cite{dosovitskiy2020image} that utilizes a large kernel for non-overlapping convolutions, our convolutional embedding uses a series of small kernels for overlapping convolutions.} 
\vspace{-10pt}\label{embedding}
\end{figure}
To alleviate this, we propose to introduce early convolutions prior to cost aggregation with transformers to control the number of tokens to reduce the costs and additionally inject some convolutional inductive bias to enhance the subsequent cost aggregation, which results in an apparent performance boost. Moreover, to reduce the computational loads from QKV projection and FFN, we introduce a novel transformer architecture for efficient cost aggregation that reformulates the two components by including appearance embedding inside the QKV projection and introducing convolutions to replace linear projections, which not only results in a costs reduction as intended but also an apparent performance boost. Thanks to the reduced costs, we are permitted to process the inputs of higher resolutions and manage to build a hierarchical architecture. Moreover, we deviate from arbitrarily selecting different combinations of feature maps~\cite{min2019hyperpixel} as was done in CATs, but rather use all the feature maps to exploit richer semantic representations. With these combined, we introduce an extension of CATs, namely CATs++.

\subsection{Feature Extraction and Cost Computation}
To exploit rich semantics present at different feature levels, we also leverage multi-level features as done in CATs and~\cite{lee2019sfnet,min2019hyperpixel,min2020learning,min2021efficient}. However, taking a slightly different approach, we use all the intermediate feature maps as done in~\cite{min2021hypercorrelation}. Specifically, similar to CATs, 
we produce a sequence of $L$ feature maps, but unlike how CATs selected a specific number of feature maps for each dataset, which is an extra burden, CATs++ use all the feature maps. In this way, not only we do not need to select different combinations of feature maps for each dataset we use, but also we can exploit richer semantics by using all the intermediate feature maps.

For the cost computation, following~\cite{min2021hypercorrelation}, given a pair of feature maps, $D^l_{s}$ and $D^l_{t}$, we compute a correlation map using the inner product between $l$-2 normalized features. We then collect correlation maps of the same spatial sizes  and denote the subset as $ \{\mathcal{C}^l\}_{l\in \mathcal{L}^{q}}$, where $\mathcal{L}^{q}$ is a subset of CNN layer indices $\{1,...,L\}$ at some pyramid layer ${q}$. Lastly, we concatenate all the colleted $\{\mathcal{C}^l\}_{l\in \mathcal{L}^{q}}$ to obtain a hypercorrelation $\mathcal{C}^{q} \in \mathbb{R}^{h_{s}w_{s} \times h_{t}w_{t} \times |\mathcal{L}^{q}|}$, where $h_{s}, w_{s}$ and $h_{t},  w_{t}$ are height and width of feature maps of source and target images, respectively.

\begin{figure*}[t]
\centering
\renewcommand{\thesubfigure}{}
\subfigure[]
{\includegraphics[width=0.123\textwidth]{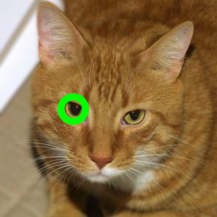}}\hfill
\subfigure[]
{\includegraphics[width=0.123\textwidth]{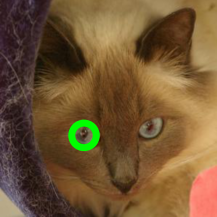}}\hfill
\subfigure[]
{\includegraphics[width=0.123\textwidth]{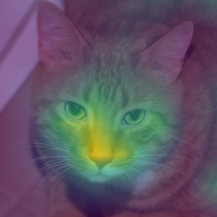}}\hfill
\subfigure[]
{\includegraphics[width=0.123\textwidth]{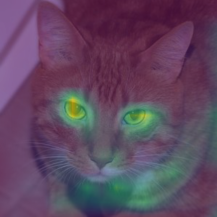}}\hfill
\subfigure[]
{\includegraphics[width=0.123\textwidth]{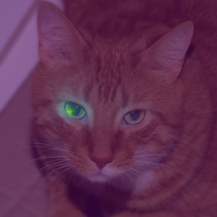}}\hfill
\subfigure[]
{\includegraphics[width=0.123\textwidth]{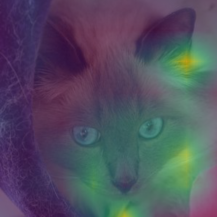}}\hfill
\subfigure[]
{\includegraphics[width=0.123\textwidth]{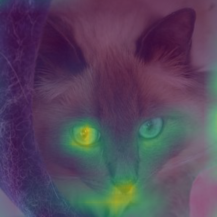}}\hfill
\subfigure[]
{\includegraphics[width=0.123\textwidth]{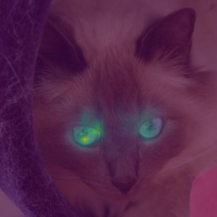}}\hfill\\
\vspace{-20.5pt}
\subfigure[(a) Source]
{\includegraphics[width=0.123\textwidth]{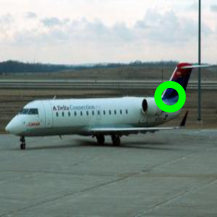}}\hfill
\subfigure[(b) Target]
{\includegraphics[width=0.123\textwidth]{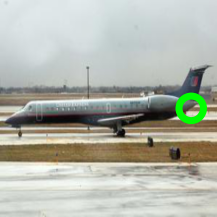}}\hfill
\subfigure[(c) Source $q=5$]
{\includegraphics[width=0.123\textwidth]{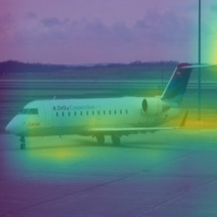}}\hfill
\subfigure[(d) Source $q=4$]
{\includegraphics[width=0.123\textwidth]{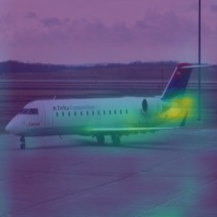}}\hfill
\subfigure[(e) Source $q=3$]
{\includegraphics[width=0.123\textwidth]{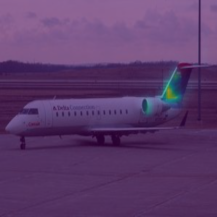}}\hfill
\subfigure[(f) Target $q=5$]
{\includegraphics[width=0.123\textwidth]{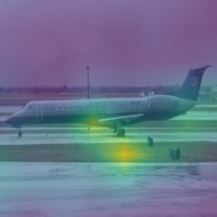}}\hfill
\subfigure[(g) Target $q=4$]
{\includegraphics[width=0.123\textwidth]{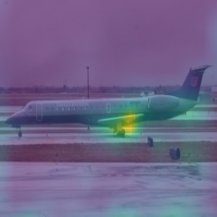}}\hfill
\subfigure[(h) Target $q=3$]
{\includegraphics[width=0.123\textwidth]{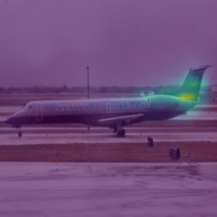}}\hfill\\
\vspace{-10pt}
\caption{\textbf{Visualization of self-attention.} Note that CATs++ utilizes hierarchical semantic representations, which the finer level self-attention is shown in the smaller region, and higher $q$ denotes coarser level.   }
\label{fig:attention}\vspace{-10pt}
\end{figure*} 
\subsection{Convolutional Embedding Module}
Without a means to control the high computational costs induced by processing the hypercorrelation $\mathcal{C}^{q}$, the amount of time, resources, and memory required would not be negligible. Considering the quadratic complexity of a standard transformer~\cite{vaswani2017attention}, simply treating all the spatial dimensions as tokens seems unwise. A better approach and the most straightforward approach, perhaps, would be to reduce the number of tokens by adopting 4D spatial pooling across source and target dimensions. However, this strategy risks losing some information, which we ought to avoid and will be verified in Section 5.4.1. {An alternative option could be to patchify the hypercorrelation by splitting it into non-overlapping tensors and embed with a large learnable kernel as done in ViT~\cite{dosovitskiy2020image}. Na{\"\i}ve implementation to this could be extending 2D patch embedding to 4D as shown in Fig.~\ref{embedding} (a). However, as claimed in~\cite{hong2022cost}, non-overlapping operations only provides limited inductive bias, which only the relatively lower translation equivariance is achieved without replacing such operations. Moreover, disregarding window boundaries due to non-overlapping kernels may hurt overall performance due to discontinuity.}

{To compensate for the issues, analogously to VAT~\cite{hong2022cost},} we use a small early convolution layer to process the correlation map prior to transformer aggregation. Given a hypercorrelation $\mathcal{C}^{q}$, we consider receptive fields of 4D window, \textit{i.e.,} $m\times m\times m\times m$, and build a tensor $\mathcal{M}^{q} \in \mathbb{R}^{\hat{h}_{s}\times \hat{w}_{s} \times \hat{h}_{t} \times \hat{w}_{t} \times |\mathcal{L}^{q}|}$, where $\hat{h}$ and $\hat{w}$ are the processed sizes as illustrated in Fig.~\ref{embedding} (b). {Note that we reduce both spatial dimensions within hypercorrelation for computational efficiency, while VAT~\cite{hong2022cost} preserves its spatial dimension of query image to obtain a fine-grained segmentation map.} During this operation, we locally mix the multi-level similarity vector at each 4D position, which acts as a local inter-correlation aggregation. In this way, we enhance the subsequent cost aggregation with transformers by injecting some convolutional bias to mitigate the above issues and allowing both local and global aggregation. 


\subsection{Efficient Transformer Aggregator} 
Although the convolutional embedding module adjusts the number of tokens and feature dimensions to some extent, directly applying standard transformers~\cite{vaswani2017attention} is still challenging. This is due to what some works~\cite{dosovitskiy2020image,vaswani2017attention,marin2021token} stated that the standard self-attention mechanism has quadratic complexity with respect to the number of tokens, and QKV projections and FFN have quadratic complexity with respect to the feature dimensionality. To address this, recently, numerous works~\cite{kitaev2020reformer,wang2020linformer,katharopoulos2020transformers,wu2021fastformer,yang2021focal} have proposed to reduce the computational burden by introducing an efficient transformer or self-attention, \textit{i.e.,} token pruning~\cite{rao2021dynamicvit}, linear self-attention~\cite{wang2020linformer} and locality-sensitive hashing~\cite{kitaev2020reformer}. However, although these methods effectively alleviate the computational burden by aiming to either control the number of tokens or reduce the complexity of self-attention, in our context, QKV projections and FFN  relatively impose more computational burden as we flatten one of the spatial dimension, \textit{i.e.,} the spatial dimension of source, and level dimension to treat them as feature dimension for a token. This has been relatively underexplored~\cite{jaszczur2021sparse}, and we propose a novel design specialized in matching cost aggregation that addresses the aforementioned issues.

\begin{figure*}[!t]
    \centering
    \renewcommand{\thesubfigure}{}
    \subfigure[]
	{\includegraphics[width=0.247\linewidth]{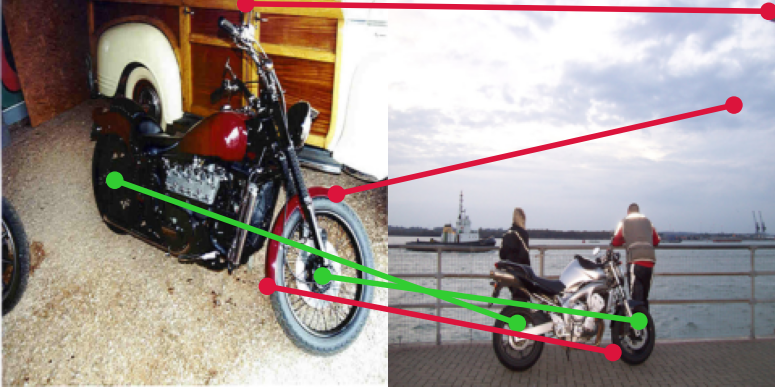}}\hfill
    \subfigure[]
	{\includegraphics[width=0.247\linewidth]{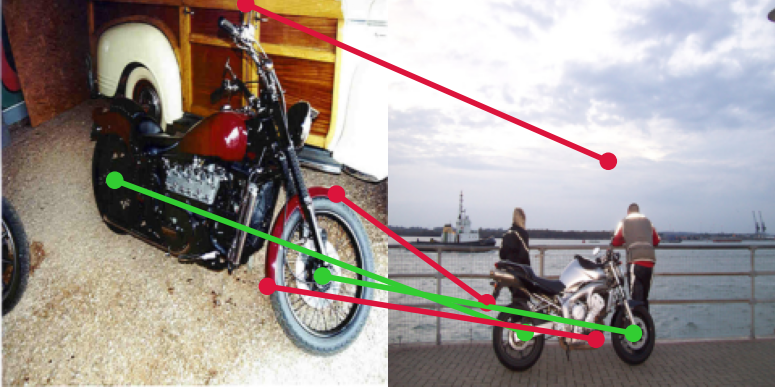}}\hfill
    \subfigure[]
	{\includegraphics[width=0.247\linewidth]{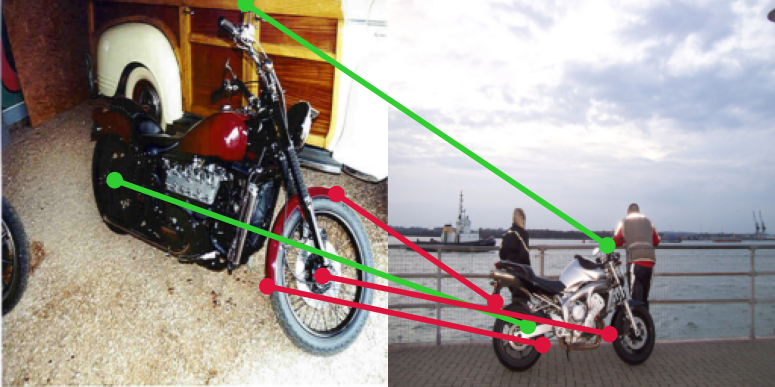}}\hfill
    \subfigure[]
	{\includegraphics[width=0.247\linewidth]{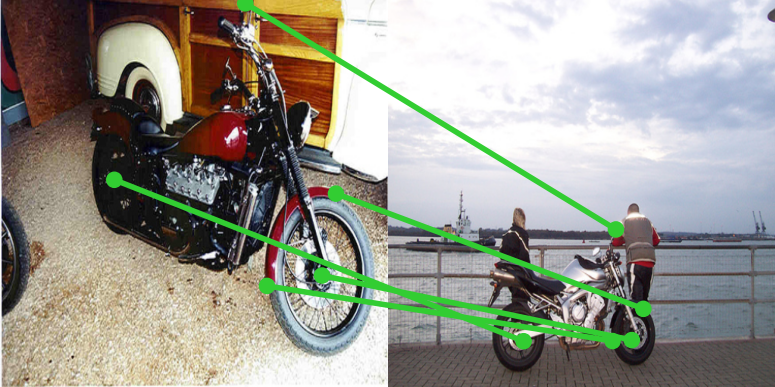}}\hfill\\
	\vspace{-20.5pt}
    \subfigure[]
	{\includegraphics[width=0.247\linewidth]{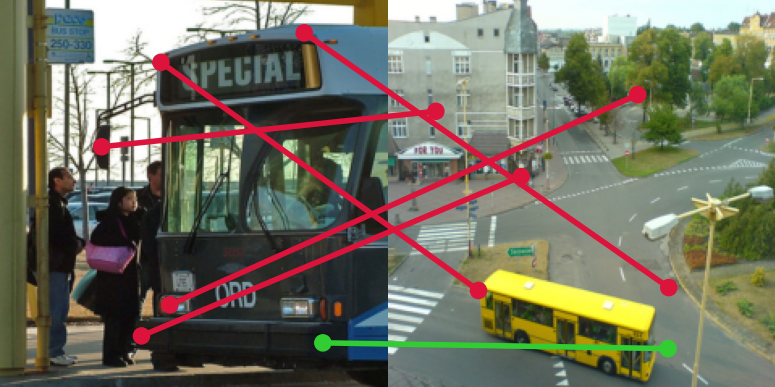}}\hfill
    \subfigure[]
	{\includegraphics[width=0.247\linewidth]{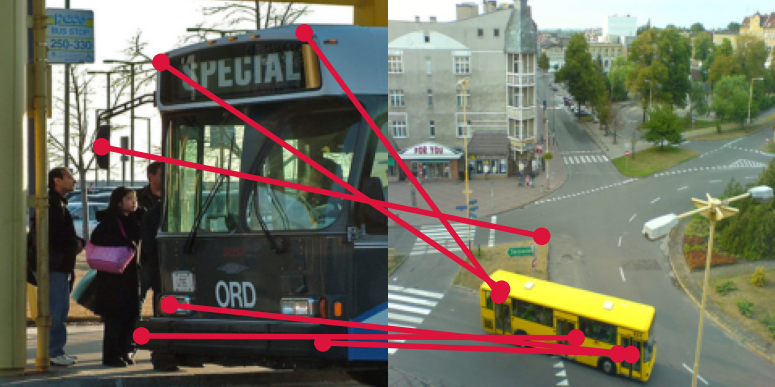}}\hfill
    \subfigure[]
	{\includegraphics[width=0.247\linewidth]{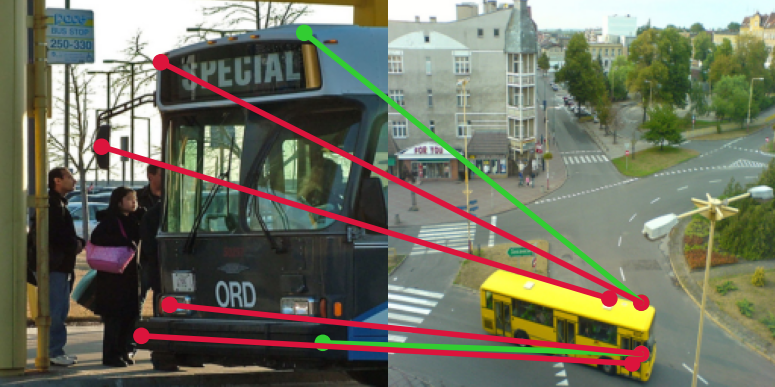}}\hfill
    \subfigure[]
	{\includegraphics[width=0.247\linewidth]{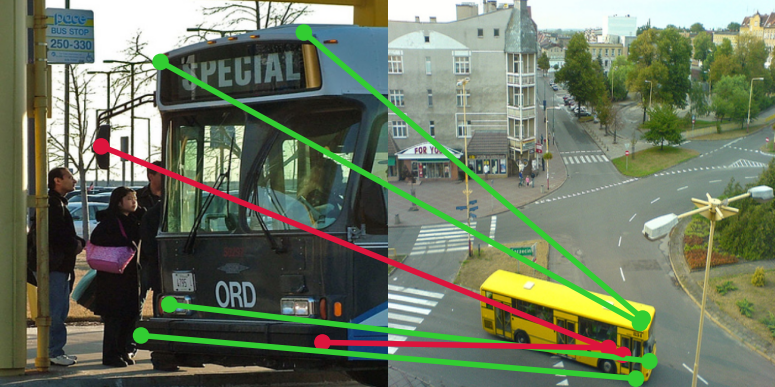}}\hfill\\
	\vspace{-20.5pt}
	\subfigure[]
	{\includegraphics[width=0.247\linewidth]{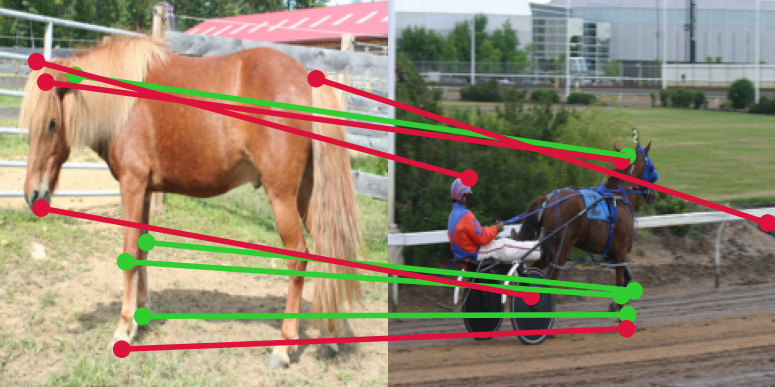}}\hfill
    \subfigure[]
	{\includegraphics[width=0.247\linewidth]{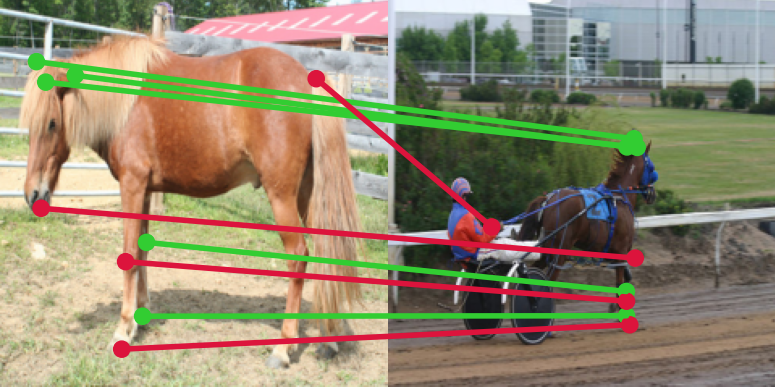}}\hfill
    \subfigure[]
	{\includegraphics[width=0.247\linewidth]{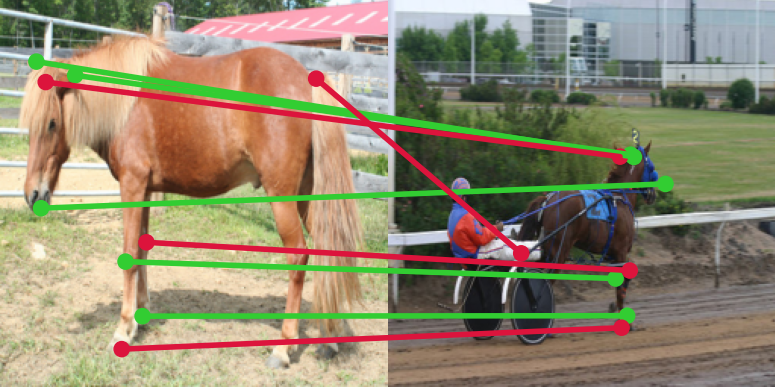}}\hfill
    \subfigure[]
	{\includegraphics[width=0.247\linewidth]{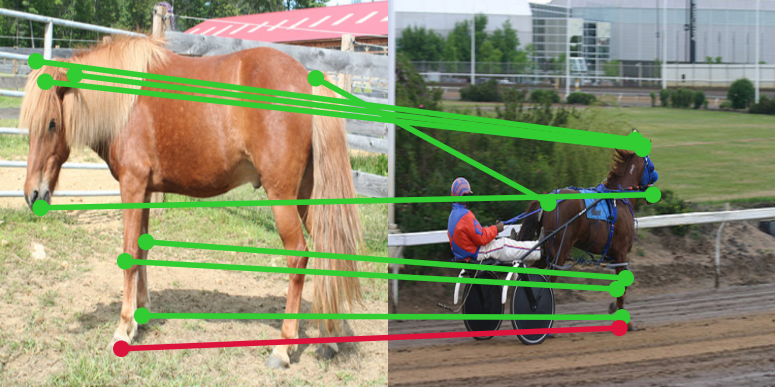}}\hfill\\
	\vspace{-20.5pt}
    \subfigure[(a) DHPF]
	{\includegraphics[width=0.247\linewidth]{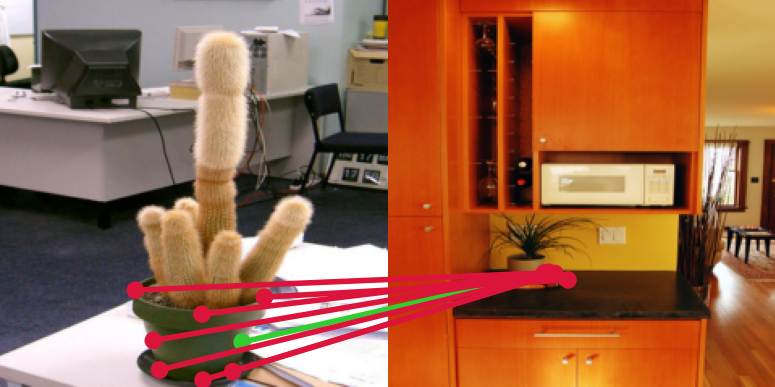}}\hfill
    \subfigure[(b) CHM]
	{\includegraphics[width=0.247\linewidth]{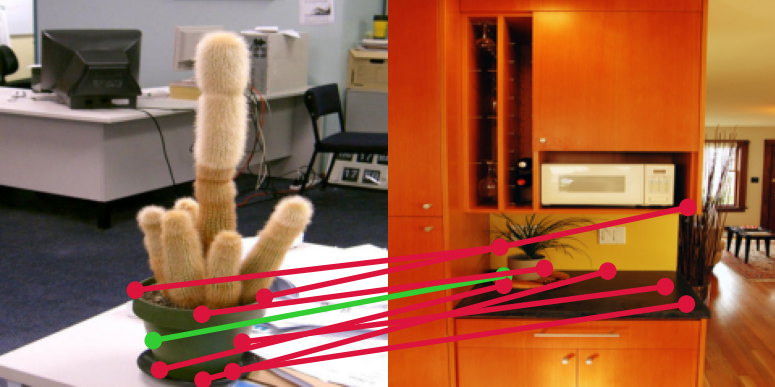}}\hfill
    \subfigure[(c) CATs]
	{\includegraphics[width=0.247\linewidth]{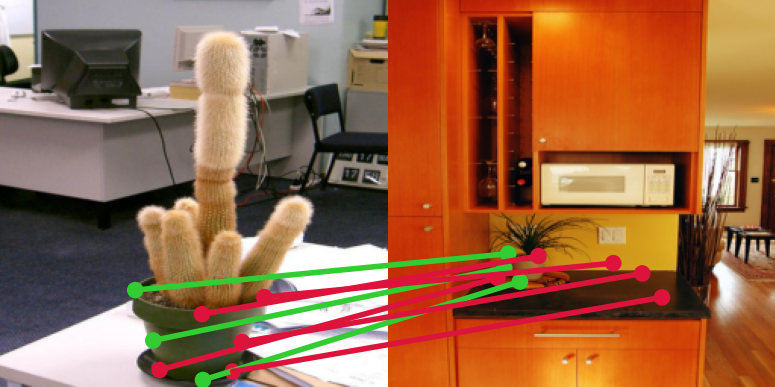}}\hfill
    \subfigure[(d) CATs++]
	{\includegraphics[width=0.247\linewidth]{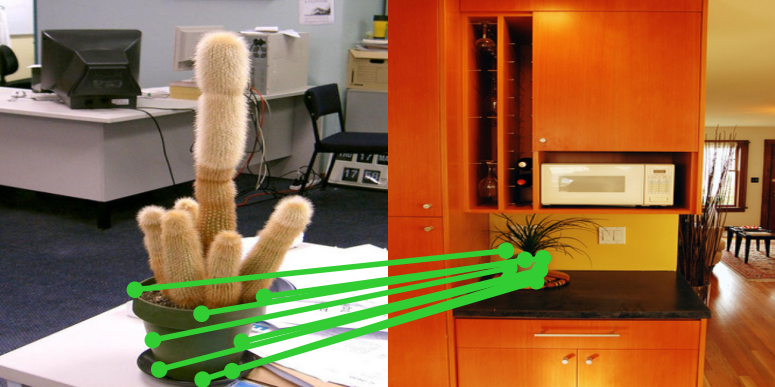}}\hfill\\
	\vspace{-10pt}
    \caption{\textbf{Qualitative results on SPair-71k~\cite{min2019spair}:} keypoints transfer results by DHPF~\cite{min2020learning}, CHM~\cite{min2021convolutional}, CATs, and CATs++. Note that the green and red line denote correct and wrong prediction, respectively, with respect to the ground-truth.}\label{fig:spair_quali}\vspace{-10pt}
  
\end{figure*}

\subsubsection{Affinity-Aware Convolutional Projection}
As illustrated in Fig.~\ref{efficient}, the proposed QKV projection differs from that of CATs that follows the standard approach in that it uses convolutions to first process high resolution input to control the subsequent computation by reducing the spatial dimension from $hw$ to $hw/s^2$, where $s$ denotes a stride size. By introducing convolutions prior to linear projection, not only we provide efficiency by allowing feature dimension controllable, but also strengthen the power to consider locality. While~\cite{wu2021cvt} tries a similar approach by applying depth-wise separable convolutions to reduce the number of tokens, here we aim to reduce the feature dimension. Subsequently, we concatenate appearance embedding to the intermediate outputs of query and queue projections, introducing appearance affinity inside the transformer architecture.  It should be noted that linear projection is included only within the Query and Key projections as we aim to not only utilize and preserve the spatial resolution of the raw correlation map but also to further reduce the computation as we find that it is sufficient to only include appearance affinity information within the attention score computation $QK^T$ rather than within the $(QK^T)\cdot V$ computation in order to aggregate both the appearance affinity along with the matching scores, which the QK projection is responsible for. Then we further reduce the feature dimension for $Q$ and $K$ projection by feeding them into linear projection, reducing self-attention computational loads. In this way,  we benefit from the reduced computation, eliminating one-third of computational loads of value projection, had the linear projection been included.

Concretely, the process can be expressed as follows:
\begin{equation}
\begin{split}
   &Q^{q} = \mathcal{P}^{q}_Q([\mathrm{LN}(\mathcal{Q}^{q}_Q(\mathrm{LN}(\mathcal{M}^q))), \mathcal{P}^{q}(D^q)]) + E_\mathrm{pos}^{q}, \\
   &K^{q} = \mathcal{P}^{q}_K([\mathrm{LN}(\mathcal{Q}^{q}_K(\mathrm{LN}(\mathcal{M}^q))), \mathcal{P}^{q}(D^q)]) + E_\mathrm{pos}^{q},\\
   &V^{q} = \mathrm{LN}(\mathcal{Q}^{q}_V(\mathrm{LN}(\mathcal{M}^q))),
   \end{split}
\end{equation}
where $\mathcal{Q}$ denotes convolutional projection; $E_\mathrm{pos}^{q}$ denotes positional embedding at $q$-th layer. Note that projections are performed at each $q$-th layer.

\subsubsection{Volumetric Convolutional Feed-Forward}
Given outputs of QKV projections, we follow the standard self-attention computation to obtain $\hat{z}$ as in Eq.~\ref{selfatt}. Note that we find it relatively less influential to change it to other efficient self-attention computation methods~\cite{wang2020linformer,kitaev2020reformer,xiong2021nystromformer}, and we thus use standard self-attention~\cite{dosovitskiy2020image} for simplicity. Subsequent to self-attention computation, in order to model additional locality bias, reduce the number of parameters for better generalization and provide more memory efficiency, we replace linear transformations and ReLU activation with 4D convolutions and GELU, respectively, as shown in Fig.~\ref{efficient}. Moreover, it should be noted that flattening the correlation volume inevitably requires a linear projection with a large feature dimension, which results in a significant computation load. This provides efficiency benefits to using convolution over a large linear projection layer. The process is defined as:
\begin{equation}
\begin{split}
    &z = \hat{z} + \mathcal{M}^q,\\
    &x = \mathrm{LN}(\mathrm{z}),\\
    &\hat{x} = \mathcal{Q}_{\mathrm{FFN}}^2(\mathrm{GELU}(\mathcal{Q}_{\mathrm{FFN}}^1(x))),\\
    &y = \hat{x} + z,
\end{split}
\end{equation}
{where $\mathcal{Q}_{\mathrm{FFN}}^1$ and $\mathcal{Q}_{\mathrm{FFN}}^2$ are the first and the second convolutional projections within the feed-forward network, respectively.} We find that this simple change brings a surprisingly apparent performance boost when combined with our proposed architecture, which will be discussed in Section 5.4.5.
\subsection{Cost Aggregation with Pyramidal Processing}
Combining all the proposed components of CATs++, we manage to reduce the costs and process higher resolution inputs with pyramidal architecture.  Specifically, analogous to~\cite{sun2018pwc,jeon2018parn,melekhov2019dgc,truong2020glu,Hong_2021_ICCV,min2021hypercorrelation} and inspired by several works~\cite{rocco2018neighbourhood,truong2020glu,Hong_2021_ICCV,min2021hypercorrelation}, we also utilize the coarse-to-fine approach through a pyramidal processing as illustrated in Fig.~\ref{fig:CATs++} to allow matching scores processed at previous levels to guide the aggregation process at finer levels by employing residual connections. In this way, we enforce the networks to focus on learning complementary matching scores~\cite{zhao2021multi} as well as let the coarse level aggregation guide the finer level aggregation. 

Overall, we define the whole process of our pyramidal cost aggregation as follows:
\begin{equation}
\begin{split}
    &\hat{\mathcal{C}}^{q} = \mathcal{T}^{q}_{+}(\mathcal{M}^q, \mathcal{P}^q(D^q)) + \mathcal{T}^{q}_{+}((\mathcal{M}^q)^T, \mathcal{P}^q(D^q)), \\
    &{\mathcal{M }}^{q-1} \leftarrow \mathrm{up}(\hat{\mathcal{C}}^{q}) + \mathcal{M}^{q-1},
\end{split}
\end{equation}
where $\mathcal{T}_{+}$ denotes efficient transformer aggregator and $\mathrm{up}(\cdot)$ denotes bilinear upsampling.

\subsection{Comparison to Concurrent Work}

{Recently, a concurrent work, Volumetric Aggregation with Transformers (VAT)~\cite{hong2022cost}, also proposed a Transformer-based architecture to tackle few-shot segmentation task. Although VAT shares a similar high-level idea, which is to combine convolutions and Transformer, and evaluates its effectiveness on both few-shot segmentation and semantic correspondence tasks, there exist apparent differences, which we summarize in the following.}

{First of all, the motivations behind the designs of each architecture differ as their main target tasks are different, \textit{i.e.,} few-shot segmentation and semantic correspondence. Although they may share similar challenges, \textit{e.g.,} background clutters and intra-class variations, at high-level, VAT~\cite{hong2022cost} designs its architecture to enhance its generalization power to handle new classes beyond training samples tailored for few-shot segmentation, while CATs++ redesigns the components of the standard Transformer for efficient computations tailored for semantic correspondence.}

{In this regard, VAT~\cite{hong2022cost} and our CATs++ formulated different transformer aggregators. It should be noted that the components prior to transformer aggregator, \textit{i.e.,} hypercorrelation and early convolutions, are the designs that both architectures share, which means that the subsequent modules lead to notable variations in effectiveness and efficiency. VAT~\cite{hong2022cost} adopts Swin Transformer~\cite{liu2021swin} as its transformer aggregator and extends to handle 4D cost volumes with 4D local windows, while CATs++ redesigns the standard transformer to a more efficient version. To address the complexity, VAT~\cite{hong2022cost} adopts shifting window multi-head self-attention~\cite{liu2021swin} that is designed to reduce the computational complexity of a global multi-head self-attention module. Unlike this, a key contribution that differentiates the proposed method to other Transformer-based architecture is that CATs++ reformulates the standard QKV projections and FFN, and proposes  affinity-aware convolutional projections and volumetric convolutional feed-forward network, aiming to reduce the memory and computational complexity that are typically high when the input is high dimensional correlation maps.}

{More importantly, we argue that the relative positioning bias that Swin Transformer~\cite{liu2021swin} include is what makes it excel at predicting segmentation map for unseen classes. We observe that relative positioning bias allows greater generalization power, which will be evidenced in Table~\ref{tab:costaggregator} in experiments, which was also observed in Swin Transformer~\cite{liu2021swin}, and this led to performance improvement for few-shot segmentation benchmark. However, for the semantic matching task, where generalization power has a relatively smaller influence to performance than to few-shot segmentation, we observe CATs++ outperforms VAT. This can be explained by the special design of CATs++ that integrates convolutions and appearance embedding with the proposed transformer aggregator, which not only allows an improved efficiency, but also a clear difference in architecture that leads to an apparent performance difference to VAT~\cite{hong2022cost} for semantic correspondence task.}


\section{Experiments}

\subsection{Implementation Details}
For the backbone feature extractor, we use ResNet-101~\cite{he2016deep} pre-trained on ImageNet~\cite{deng2009imagenet} for both CATs and CATs++, while other backbone features can also be used. For the hyperparameters of the transformer encoder, we set the number of encoders as 1 for all the transformer aggregators in both CATs and CATs++ based on the ablation study, which we chose to set it as 1 for simplicity. For CATs, we resize the spatial size of the input image pairs to 256$\times$256 and a sequence of selected features is resized to $h$ = 16 and $w$ = 16, while we resize image pairs to 512$\times$512 for CATs++. For pyramidal processing in CATs++, we employ pyramidal layers of $q=3,4,5$: from $\mathrm{conv3}\_x$ to $\mathrm{conv5}\_x$,  similar to~\cite{min2021hypercorrelation}. AdamW~\cite{loshchilov2017decoupled} optimizer with an initial learning rate of $3\mathrm{e}-5$ for the aggregators and $3\mathrm{e}-6$ for the backbone features are used, which we gradually decrease during training. To apply data augmentation~\cite{cubuk2020randaugment,buslaev2020albumentations} with predetermined probabilities to input images at random. Specifically, 50$\%$ of the time, we randomly crop the input image, and independently for each augmentation function used in~\cite{cubuk2020randaugment}, we set the probability for applying the augmentation as 20$\%$. We implemented our network using PyTorch~\cite{paszke2017automatic}.

\subsection{Experimental Settings}
In this section, we conduct comprehensive experiments for semantic correspondence, by evaluating our approach through comparisons to state-of-the-art methods including CNNGeo~\cite{rocco2017convolutional}, A2Net~\cite{paul2018attentive}, WeakAlign~\cite{rocco2018end}, NC-Net~\cite{rocco2018neighbourhood},
RTNs~\cite{kim2018recurrent}, SFNet~\cite{lee2019sfnet}, HPF~\cite{min2019hyperpixel}, DCC-Net~\cite{huang2019dynamic}, ANC-Net~\cite{li2020correspondence},  DHPF~\cite{min2020learning}, SCOT~\cite{liu2020semantic}, GSF~\cite{jeon2020guided}, and CHM~\cite{min2021convolutional}, CATs~\cite{cho2021semantic}, PMNC~\cite{lee2021patchmatch}, MMNet~\cite{zhao2021multi}, and CHMNet~\cite{min2021efficient}.

\subsubsection{Datasets}
We consider SPair-71k~\cite{min2019spair}  which provides a total of 70,958 image pairs with extreme and diverse viewpoints, scale variations, and rich annotations for each image pair, e.g., keypoints, scale difference, truncation and occlusion difference, and clear data split. Previously, for semantic matching, most of the datasets are limited to a small quantity with similar viewpoints and scales~\cite{ham2016proposal,ham2017proposal}. As our network relies on transformer which requires a large number of data for training, SPair-71k~\cite{min2019spair} makes the use of transformer in our model feasible. 
We also consider PF-PASCAL~\cite{ham2017proposal} containing 1,351 image pairs from 20 categories and PF-WILLOW~\cite{ham2016proposal} containing 900 image pairs from 4 categories, each dataset providing corresponding ground-truth annotations.

\begin{table*}[!t]
    \begin{center}
    \caption{\textbf{Quantitative evaluation on standard benchmarks~\cite{min2019spair,ham2016proposal,ham2017proposal}.} Higher PCK is better. The best results are in bold, and the second best results are underlined. CATs$\dagger$ (or CATs++$^\dagger$) denotes CATs (or CATs++) without fine-tuning feature backbone. Note that ResNet-101 is used as a backbone feature extractor as a default setting unless stated otherwise, \textit{i.e.,} MMNet~\cite{zhao2021multi}.
    ~\emph{Feat.-level: Feature-level, FT. feat.: Fine-tune feature.} 
    }\label{tab:main_table}\vspace{-5pt}
    \scalebox{0.9}{
    \begin{tabular}{l|c|c|c|c|c|ccc|cc|cc}
            \hlinewd{0.8pt}
             \multirow{3}{*}{Methods}&\multirow{3}{*}{Data Aug.}&\multirow{3}{*}{Feat.-level} &\multirow{3}{*}{FT. feat.} & \multirow{3}{*}{Cost Aggregation}& SPair-71k~\cite{min2019spair} & \multicolumn{3}{c|}{PF-PASCAL~\cite{ham2017proposal}} & \multicolumn{4}{c}{PF-WILLOW~\cite{ham2016proposal}} \\
          
              &&&& &PCK @ $\alpha_{\text{bbox}}$ & \multicolumn{3}{c|}{PCK @ $\alpha_{\text{img}}$}  & \multicolumn{2}{c}{PCK @ $\alpha_{\text{bbox}}$} & \multicolumn{2}{c}{PCK @ $\alpha_{\text{bbox-kp}}$} \\ 
        
              && & & &0.1 & 0.05 & 0.1 & 0.15 & 0.05 & 0.1 & 0.05 &0.1 \\ 
             \midrule
             WTA &\xmark&Single &\xmark&-& 25.7&35.2&53.3&62.8&30.1&52.9&24.7&46.9\\\midrule
             CNNGeo~\cite{rocco2017convolutional}&{\xmark} &Single&\xmark&- &20.6 &41.0 &69.5 &80.4 &- &- &36.9&69.2\\
             A2Net~\cite{paul2018attentive}&\xmark &Single&\xmark&- &22.3 &42.8 &70.8 &83.3 &- &- &36.3&68.8\\
              WeakAlign~\cite{rocco2018end}&\xmark &Single&\xmark&- &20.9 &49.0 &74.8 &84.0 &- &- &37.0&70.2\\
              RTNs~\cite{kim2018recurrent}&\xmark &Single &\xmark & - &25.7 &55.2 &75.9 &85.2 &- &- &41.3&71.9\\ 
              SFNet~\cite{lee2019sfnet}&\xmark &Multi &\xmark & - &- &53.6 &81.9 &90.6 &- &- &46.3&74.0\\
              MMNet-{FCN}~\cite{zhao2021multi}&\xmark &Multi &\cmark & - & 50.4&81.1 & 91.6& 95.9&- &- &-&-\\
        
             \midrule
             NC-Net~\cite{rocco2018neighbourhood}&{\xmark} &Single &\cmark &4D Conv. &20.1 &54.3 &78.9 &86.0 &- &- &33.8&67.0\\

             DCC-Net~\cite{huang2019dynamic}&\xmark &Single &\xmark &4D Conv. &- &55.6 &82.3 &90.5 &- &- &43.6&73.8\\
             
             HPF~\cite{min2019hyperpixel}& \xmark&Multi &- &RHM&28.2 &60.1 &84.8 &92.7 &- &- &45.9&74.4\\
             GSF~\cite{jeon2020guided}&\xmark &Multi &\xmark &2D Conv. &36.1&65.6 &87.8 &95.9 &- &- &\textbf{49.1}&\textbf{78.7}\\
             ANC-Net~\cite{li2020correspondence}&\xmark &Single &\xmark &4D Conv. &- &- &86.1 &- &- &- &-&-\\
             DHPF~\cite{min2020learning}& \xmark&Multi &\xmark&RHM &37.3 &75.7 &90.7 &95.0 &49.5 &77.6 &- &71.0\\
             SCOT~\cite{liu2020semantic}& \xmark&Multi &- &OT-RHM &35.6 &63.1 &85.4 &92.7 &- &- &\underline{47.8}&\underline{76.0}\\
             CHM~\cite{min2021convolutional}&\xmark &Single &\cmark & 6D Conv. &46.3 &80.1 &91.6 &94.9 &52.7 &{79.4} &-&69.6\\
             PMNC~\cite{lee2021patchmatch}&\xmark &Multi &\cmark & 4D Conv. & 50.4& \underline{82.4}& 90.6& -& -&- &-&-\\

             CHMNet~\cite{min2021efficient}&\cmark &Multi &\cmark & 6D Conv. & {51.3}& 81.3& \underline{92.9}& - &53.8& 79.3&-&69.3\\
             {VAT~\cite{hong2022cost}}&{\cmark} &{Multi} &{\cmark} & {4D Conv. +  Transformer} &{\underline{55.5}} &{78.2} &{92.3} &{96.2}  &{52.8}&{\textbf{81.6}} &{-}&{-}\\\midrule
           
             CATs$\dagger$&\cmark &Multi &\xmark &Transformer&42.4 &67.5 &89.1 &94.9 &46.6 &75.6 & 37.3&65.7\\
             CATs&\cmark &Multi &\cmark &Transformer&49.9 &75.4 &92.6 &\underline{96.4} &50.3 &79.2 & 40.7&69.0\\\midrule
            
             CATs++$\dagger$&\cmark &Multi &\xmark &{4D Conv. + Transformer}&{50.0} &73.0 &89.6 &95.0 &\underline{54.1} &78.5 &44.5&69.9\\
     
             CATs++&\cmark &Multi &\cmark &{4D Conv. + Transformer}&\textbf{59.8} & \textbf{84.9}&\textbf{93.8} &\textbf{96.8} &\textbf{56.7} &\underline{81.2} &47.0&72.6\\
             
            \hlinewd{0.8pt}
    \end{tabular}
    }
    \end{center}\vspace{-5pt}
\end{table*}
\subsubsection{Evaluation Metric}
For evaluation on SPair-71k~\cite{min2019spair}, PF-PASCAL~\cite{ham2017proposal}, and PF-WILLOW~\cite{ham2016proposal}, we employ a percentage of correct keypoints (PCK), computed as the ratio of estimated keypoints within the threshold from ground-truths to the total number of keypoints. Given predicted keypoint $k_\mathrm{pred}$ and ground-truth keypoint $k_\mathrm{GT}$, we count the number of predicted keypoints that satisfy the following condition: $d( k_\mathrm{pred},k_\mathrm{GT}) \leq \alpha \cdot \mathrm{max}(H,W)$, where $d(\,\cdot\,)$ denotes Euclidean distance; $\alpha$ denotes a threshold. We evaluate on PF-PASCAL with $\alpha_\mathrm{img}$, SPair-71k, and PF-WILLOW with $\alpha_\mathrm{bbox}$. $H$ and $W$ denote height and width of the object bounding box or the entire image, respectively. As stated in~\cite{min2021efficient}, we additionally report results of $\alpha_\text{bbox-kp}$ for PF-WILLOW~\cite{ham2016proposal} for a fair comparison.

\subsection{Semantic Correspondence Results}\label{5.3}
For a fair comparison, we follow the evaluation protocol of~\cite{min2019hyperpixel} for SPair-71k~\cite{min2019spair}, in which our network is trained on the training split and evaluated on the test split. Similarly, for PF-PASCAL~\cite{ham2017proposal} and PF-WILLOW~\cite{ham2016proposal}, following the common evaluation protocol of~\cite{han2017scnet,kim2018recurrent,huang2019dynamic,min2019hyperpixel,min2020learning}, we train our network on the training split of PF-PASCAL~\cite{ham2017proposal} and then evaluate on the test split of PF-PASCAL~\cite{ham2017proposal} and PF-WILLOW~\cite{ham2016proposal}. All the results of other methods are reported under the identical setting.

\begin{table}[]
    \centering
    \caption{\textbf{Ablation study of CATs.}}
    \label{tab:ablearch}\vspace{-5pt}
    \scalebox{1}{
    \begin{tabular}{cl|cc}
    \hlinewd{0.8pt}
         &\multirow{4}{*}{Components} &\multicolumn{2}{c}{SPair-71k}  \\
        &&\multicolumn{2}{c}{$\alpha_{\mathrm{bbox}}$ = 0.1}\\\cline{3-4}
        &&\multicolumn{2}{c}{FT. feat.}\\\cline{3-4}
        &&\xmark&\cmark\\
        \midrule
       \textbf{(I)} &Baseline  &26.8 &{46.7}\\
        \textbf{(II)} &+ Appearance Modelling  &33.5 &{46.3}\\
        \textbf{(III)} &+ Multi-level Aggregation &35.9&{47.0} \\
        \textbf{(IV)} &+ Swapping Self-Attention &38.8 &{47.6}\\
        \textbf{(V)} &+ Residual Connection &42.4 &{49.9}\\\hlinewd{0.8pt}
    \end{tabular}
    }\vspace{-5pt}
    
\end{table}

\begin{table}[]
    \centering
     \caption{\textbf{Ablation study of CATs++.}}
    \label{tab:ablearch2}\vspace{-5pt}
    \scalebox{1}{
    \begin{tabular}{cl|c|c}
    \hlinewd{0.8pt}
         &\multirow{4}{*}{Components} &\multicolumn{2}{c}{SPair-71k}  \\
        &&\multicolumn{2}{c}{$\alpha_{\mathrm{bbox}}$ = 0.1} \\\cline{3-4}
        &&\multicolumn{2}{c}{FT. feat.}\\\cline{3-4}
        &&\xmark&\cmark\\
        \midrule
        \textbf{(I)} &Baseline  & 22.7 &{49.8} \\
        \textbf{(II)} &+ Hypercorrelation  &20.2 &{50.2} \\
        \textbf{(III)} &+ Convolutional Embedding  &27.9 &{54.1} \\
        \textbf{(IV)} &+ Efficient Transformer Aggregator &42.1 &{55.3}\\
        \textbf{(V)} &+ Appearance Modelling &50.0 &{59.8} \\\midrule
        \textbf{(VI)} &(\textbf{IV}) - Convolutional Embedding  &{47.4} &{57.8} \\
        \textbf{(VII)} &(\textbf{V}) - Convolutional Embedding &{48.5}&{59.1} \\
        \textbf{(VIII)} &(\textbf{V}) - Hypercorrelation &{46.7} &{58.1} \\
        \textbf{(IX)} &(\textbf{VIII}) - Convolutional Embedding &{45.6} &{55.7} \\\hlinewd{0.8pt}
    \end{tabular}
    }\vspace{-5pt}
   
\end{table}

Fig.~\ref{fig:spair_quali} visualizes qualitative results for extremely challenging image pairs. We observe that our methods are capable of suppressing noisy scores and finding accurate correspondences in cases where large scale and geometric variations are present. Table~\ref{tab:main_table} summarizes quantitative results on SPair-71k~\cite{min2019spair}, PF-PASCAL~\cite{ham2017proposal} and PF-WILLOW~\cite{ham2016proposal}. In order to ensure a fair comparison, we note whether each method leverages multi-level features and fine-tunes the backbone networks. We additionally denote the types of cost aggregation.

We first compare the proposed methods with those that do not fine-tune backbone networks. Comparing CATs$\dagger$ and CATs++$\dagger$ to other methods, we find that not only CATs$\dagger$ outperforms  DHPF~\cite{min2020learning}, by 5.1$\%p$ on SPair-71k, but also CATs++$\dagger$ further boosts the performance to achieve 50.0$\%$, which makes it beyond comparison. Interestingly, the results obtained by CATs++$\dagger$ are almost on par with the previous state-of-the-art method, CHMNet~\cite{min2021efficient}, even without fine-tuning the backbone. With the backbone networks fine-tuned, CATs shows highly competitive performance, achieving similar performance to previous state-of-the-art methods, thanks to its ability to explore global consensus and powerful data augmentation technique to fulfill the need of data-hungry transformer. It should be noted that CATs++ sets a new state-of-the-art on almost all the benchmarks, and it surpasses CHMNet by 8.5$\%p$ on SPair-71k, which clearly demonstrates its effectiveness and highlights that cost aggregation is of prime importance, and leveraging both convolutions and transformers in a right way clearly makes stronger cost aggregation. {Note that VAT~\cite{hong2022cost} also achieves highly competitive results, even surpassing the results of proposed method on PF-WILLOW~\cite{ham2016proposal} at $\alpha=0.1$. 
Nevertheless, it is shown that the proposed approach outperforms for all other benchmarks, highlighting the superiority of the proposed efficient transformer to Swin Transformer~\cite{liu2021swin} in VAT~\cite{hong2022cost} for semantic correspondence task.}

Now comparing CATs and CATs++, we observe a significant performance boost for both backbone fine-tuned and frozen. Specifically, when the backbone is frozen, 7.6$\%p$ increase is observed while there is a 9.9$\%p$ increase when backbone is fine-tuned. This clearly demonstrates the effectiveness extension of CATs. Also, we observe from the results of PF-WILLOW~\cite{ham2016proposal} that more fine details are predicted correctly as well as the generalization power is enhanced. This is confirmed by how CATs++$\dagger$ is almost on par with CATs on PF-WILLOW even though their performance gap on PF-PASCAL shows a large gap.     

Additionally, as stated in CHMNet~\cite{min2021efficient}, for a fair comparison, we report results of PCK @ $\alpha_\mathrm{bbox-kp}$ for PF-WILLOW~\cite{ham2016proposal}. It is notable that our methods generally report lower PCK on PF-WILLOW~\cite{ham2016proposal} compared to other state-of-the-art methods. We conjecture that this is due to the provision of sparsely annotated data and the use of transformers, since sparse annotations can hurt the generalization power~\cite{min2021efficient}. Moreover, it is well known that the transformers lack generalization power with a small dataset size~\cite{dosovitskiy2020image}. When we evaluate on PF-WILLOW, we infer with the model trained on the training split of PF-PASCAL~\cite{ham2017proposal} with sparse keypoint annotations, which only contains 1,351 image pairs, and as an only relatively small quantity of image pairs is available within the PF-PASCAL training split, the proposed method shows low generalization power. Note that to compensate for this issue, we provided data augmentation techniques, which clearly helped to increase the generalization power. From this, we suspect that a provision of more data could help to address the generalization issue, and perhaps the weak supervisions other than sparse keypoints could further enhance its generalization power.

\subsection{Ablation study}\label{5.4}
In this section, we show an ablation analysis to validate critical components we made to design our architecture, and provide analysis on the use of different backbone features, the depth of transformer encoder, comparison among different cost aggregators, memory/run-time, and data augmentation. We train all the variants on the training split of SPair-71k~\cite{min2019spair} when evaluating on SPair-71k, and train on PF-PASCAL~\cite{ham2017proposal} for evaluating on PF-PASCAL. We measure the PCK, and each ablative experiment is conducted under the same experimental setting for a fair comparison.

\subsubsection{Effectiveness of each component}
Table~\ref{tab:ablearch} shows the analysis of key components of CATs. We analyze four key components for this ablation study, which include appearance modeling, multi-level aggregation, swapping self-attention, and residual connection. As shown in Table~\ref{tab:ablearch2}, we also conduct an ablation study on key components of CATs++, which we consider hypercorrelation, convolutional embedding, efficient transformer aggregator, and appearance modeling. For both analyses, we evaluate on SPair-71k benchmark by progressively adding each key component. {In this experiments, we report the results for the backbone models with and without fine-tuned. Moreover, to explore how the proposed modules are co-affected, we evaluate the models equipped with different combinations of the modules, and the results are reported in Table~\ref{tab:ablearch2} from \textbf{VI} to \textbf{IX}.}

In Table~\ref{tab:ablearch}, we first start with CATs by defining the model without any of the key components as a baseline model, which simply feeds the correlation map into the self-attention layer. From \textbf{I} to \textbf{V}, we observe a consistent improvement in performance when each component is added. The results of \textbf{I} show relatively lower PCK, indicating that simply utilizing transformers does not yield high performance, which highlights the importance of each of our proposed components. Interestingly, \textbf{II} shows a large improvement in performance, which we find that the appearance modeling enables the model to refine the ambiguous or noisy matching scores and the use of both appearance and cost volume allows the transformer to relate the pairwise relationships and appearance, helping to find more accurate correspondences. Although a relatively small increase of PCK for \textbf{III}, it shows that the proposed model successfully aggregates the multi-level correlation maps with the help of intra- and inter- correlation self-attention. Furthermore, \textbf{IV} and \textbf{V} show an apparent increase by helping the training process, clearly confirming the significance of both components. 
\begin{table}[t]
    \centering
    \caption{\textbf{Component analysis of efficient transformer aggregator.} A.A.C: Affinity-aware Convolutional. V.C: Volumetric Convolutional.}
    \label{tab:efficient}\vspace{-5pt}
    \scalebox{1}{
    \begin{tabular}{l|c|c|c}
\hlinewd{0.8pt}
\multirow{2}{*}{Component}&SPair-71k &Memory & Num. of \\
&$\alpha_\mathrm{bbox}$ = 0.1& [GB] & param. \\
\midrule
 Baseline&36.7 &2.2 & 24.9M\\ 
(\textbf{I}) A.A.C QKV projection&46.6 &2.0&14.4M \\ 
(\textbf{II}) V.C feed-forward&40.0 &2.0&15.9M  \\ 
(\textbf{I}+\textbf{II}) Efficient transformer&50.0&1.9 &5.5M\\
\hlinewd{0.8pt}
\end{tabular}%
}\vspace{-5pt}
    
\end{table}

In Table~\ref{tab:ablearch2}, for CATs++, we define the baseline model that does not utilize appearance affinity, only uses feature maps of the last index at each $p$-th layer, replaces convolutional embedding with simple 4D max-pooling with linear projection, and utilizes raw correlation map without cost aggregation. In the results, we observe that each component except \textbf{II} contributes to a surprisingly large performance boost.  Especially from \textbf{III} to \textbf{IV}, the performance gap is 14.2$\%p$, which clearly demonstrates the effectiveness of the proposed efficient transformer aggregator. It should be noted that \textbf{V} is actually included within our proposed efficient transformer aggregator, which means that the proposed efficient transformer by itself brings a 22.1$\%p$ increase.  Interestingly, we observe a slight drop in performance when we attempt to use hypercorrelation as shown in \textbf{II}. We conjecture that although the use of multi-level features generally helps to find better correspondences~\cite{min2019hyperpixel,min2020learning,min2021efficient,cho2021semantic}, without a means to aggregate the costs, the use of multi-level features rather make the matching process more complex.

{ We also observe consistent improvements in performance even with backbone network fine-tuned when each module is cumulatively stacked. However, the results for \textbf{VI} and \textbf{VIII} deviate from this trend, where the PCK of \textbf{VIII} with backbone network fine-tuned is higher than that of \textbf{VI} despite lower PCK for the backbone model without fine-tuning. Also, we find that although we generally observe consistent performance improvements as we stack up the components, \textbf{II} in Table~\ref{tab:ablearch} shows a slight performance drop. This may seem trivial for only a 0.4 $\%$ drop, but  these could indicate that fine-tuning the backbone network may have had larger influence than introducing a new component. Another notable result is \textbf{VII}, which shows slightly lower PCK to the best PCK reported, indicating that although convolutional embedding can help to achieve higher performance, highly competitive results can be attained even without convolutional embedding.}

\subsubsection{Effectiveness of each component in efficient transformer aggregator}
In this ablation study, we provide an analysis of the proposed efficient transformer aggregator of CATs++ with respect to performance, memory, and number of learnable parameters. For this study, we first define the baseline model; it simply replaces the efficient transformer aggregator with standard transformer~\cite{vaswani2017attention}. We investigate the effectiveness of the proposed components by replacing each component in a standard transformer with our proposed component, as illustrated in Table~\ref{tab:efficient}. Note that for calculating both memory consumption and the number of parameters, we exclude memory and parameters of backbone networks and convolutional embedding in order to emphasize  the changes made by adopting each of our proposed efficient transformer aggregators. 

Replacing QKV projection with affinity-aware convolutional projection, we observe that the PCK improves by 9.9$\%p$. We also observe a similar increase when we replace feed-forward network with the proposed volumetric convolutional feed-forward. This apparent performance boost demonstrates their effectiveness, and this is further confirmed when two components are combined, which our efficient transformer aggregator obtains PCK of 50.0$\%p$, a 13.3$\%p$ higher results than that of the baseline model. We also report memory consumption and the number of learnable parameters to validate their efficiency. Overall, each component reduces the memory consumption by approximately 0.2 GB, which makes roughly 0.3 GB or 11$\%$ memory reduction in total. What surprised us is that the number of learnable parameters is reduced by almost 80$\%$, which may have resulted in better generalization power.  

\begin{table}[t]
    \centering
    \caption{\textbf{Ablation study of cost aggregation methods.} Note that we abandon some of our key contributions in order to implement 3D convolutions. \textit{RPB: Relative Positioning Bias.}}
    \label{tab:costaggregator}\vspace{-5pt}
    \scalebox{0.82}{
    \begin{tabular}{l|ccc}
\hlinewd{0.8pt}
\multirow{2}{*}{Aggregator} &SPair-71k &FSS-1000  &Memory \\
 &$\alpha_\mathrm{bbox}$ = 0.1&mIoU (1 shot)&[GB]\\
\midrule
 MLP&34.4 &- &2.0 \\
MLP-Mixer~\cite{tolstikhin2021mlpmixer} &{39.1}&-&2.2 \\
3D convolutions~\cite{tran2018closer} &30.6 &-&2.3 \\
Center-pivot 4D convolutions~\cite{min2021efficient} &36.9 &-&\textbf{1.7}\\
Standard Transformer~\cite{vaswani2017attention}&\underline{42.4}&\underline{80.8} &\underline{1.9}\\
{Standard Transformer~\cite{vaswani2017attention,liu2021swin} (w/ RPB)}&\textbf{{42.6}}&\textbf{{81.8}}&\underline{1.9}\\

\hlinewd{0.8pt}
\end{tabular}}\vspace{-5pt}
    
\end{table}
\subsubsection{Why transformers for cost aggregation?}
As our architecture is based on transformers, perhaps it is necessary to compare the effectiveness of transformers to other methods that are capable of relating the tokens of given inputs, \textit{i.e.,} MLP or convolutions. Especially, MLP and MLP mixer~\cite{tolstikhin2021mlpmixer} also enjoy from global receptive fields, which is similar to transformers, so we believe that this comparison is necessary to justify our choice to select transformers over others. To this end, we compare the transformer aggregator with all the other cost aggregation methods, and the results are summarized in Table~\ref{tab:costaggregator}. For this experiment, we simply replace the self-attention module with other aggregator methods without touching any of our contributions. Note that, unlike other methods, 3D convolution is not compatible with our contributions as the different structure of the spatial axis makes them infeasible to directly replace the module, which we have to abandon some of the proposed components, \textit{e.g.,} multi-level aggregation. However, we included 3D convolutions to evaluate the performance as a cost aggregator. Lastly, we adjusted the number of parameters for a fair comparison and report the memory consumption for different cost aggregators to indicate the efficiency.

From the results, using simple MLP yields surprisingly competitive results, thanks to its global cost aggregation, but its relatively poor performance highlights the importance of pairwise relationships, as MLP is only responsible for channel-mixing operation.  This is similar to convolution, as MLP also lacks the adaptability to input pixels prevents from accurate matching. Using MLP-Mixer~\cite{tolstikhin2021mlpmixer}, which includes both channel and token mixing, yields highly competitive results as well, but compared to CATs, the performance is relatively poor, which the inadaptability can be one of the reasons. Given this, we argue that the transformer better considers the pairwise relationships than MLP-based aggregators and shows its superiority to better take global context in a correlation map into account. Now comparing the results to convolution-based aggregators~\cite{tran2018closer,min2021efficient}, we observe that the performance gap is quite large for 3D convolutions. This is because we abandoned some of our contributions for this experiment. The use of center-pivot 4D convolution surprisingly performs well, but compared to transformers, the gap seems large. With such differences, CATs obtaining better performance over other methods indicates its superiority and advantage to explicitly learn pairwise relationships with global attention range, process the input as a whole, and remove locality constraint which highlights the importance of our contribution and the effectiveness of transformer aggregator. 

To compare the efficiency of different aggregators, we also note the memory consumption for each. Summarizing the results, we observe that memory consumption gap is quite large between 3D convolution and center-pivot 4D convolution, but others show relatively similar memory consumption, and transformer requires the second lowest memory.  Although center-pivot 4D convolutions shows the best efficiency thanks to its deliberate design, considering the large performance gap and relatively small memory consumption gap, we claim that the proposed transformer-based cost aggregator shows its superiority. 

{Additionally, we include the results of the model that employs Relative Positioning Bias (RPB)~\cite{liu2021swin}. In this experiment, we aim to investigate the reasoning behind the performance difference between this work and that of VAT~\cite{hong2022cost}. To this end, we include relative positioning bias within the self-attention computation and left other components untouched. In practice, we did not find notable performance changes for semantic matching benchmarks, but rather, we observed a notable performance boosts for FSS-1000~\cite{li2020fss} dataset. Note that we also observed a similar tendency for CATs++, except that we observed a slight performance drop when evaluated on SPair-71k~\cite{min2019spair}, confirming a relatively weaker influence of relative positioning bias on a semantic matching task. We thus find that the relative positioning bias can enhance the generalization power, which would explain some portion of performance gap between this work and VAT~\cite{hong2022cost}.}

\subsubsection{Does different feature backbone matter?}
As shown in Table~\ref{tab:backbone}, we explore the impact of different feature backbones on the performance on SPair-71k~\cite{min2019spair} and PF-PASCAL~\cite{ham2017proposal}. We report the results with backbone networks frozen.  For subscript \texttt{single}, we use a single-level feature, while for subscript \texttt{all}, every feature map from 12 layers is used for cost construction. For ResNet-101 subscript \texttt{multi}, we use the best layer subset provided by~\cite{min2019hyperpixel}, while we use hypercorrelation approach for subscript $\texttt{all}$.
\begin{table}[t]\centering
 \caption{\textbf{Ablation study of feature backbone of CATs.}}
    \label{tab:backbone}\vspace{-5pt}
    \scalebox{1}{

    \begin{tabular}{l|c|c}
         \hlinewd{0.8pt}
\multirow{2}{*}{Feature Backbone} &SPair-71k &PF-PASCAL\\

 &$\alpha_\mathrm{bbox}$ = 0.1 &$\alpha_\mathrm{img}$ = 0.1 \\
               \midrule
DeiT-B$_\texttt{single}$~\cite{touvron2020deit}&32.1 &76.5\\
DeiT-B$_\texttt{all}$~\cite{touvron2020deit}&38.2 &87.5\\\midrule
DINO w/ ViT-B/16$_\texttt{single}$~\cite{caron2021emerging} &39.5 &88.9 \\
DINO w/ ViT-B/16$_\texttt{all}$~\cite{caron2021emerging} &42.0 &88.9 \\\midrule

ResNet-101$_\texttt{single}$~\cite{he2016deep} & 37.4 &87.3  \\
ResNet-101$_\texttt{multi}$~\cite{he2016deep} &42.4 &\underline{89.1}\\\midrule
ViT w/ CLIP$_\texttt{single}$~\cite{radford2021learning} & 36.8 & 85.3
\\
ViT w/ CLIP$_\texttt{all}$~\cite{radford2021learning} & \textbf{47.9}& 89.0\\\midrule
ResNet-101 w/ CLIP$_\texttt{single}$~\cite{radford2021learning} &30.8  & 77.2
\\
ResNet-101 w/ CLIP$_\texttt{all}$~\cite{radford2021learning} &\underline{45.3} &\textbf{89.4}\\ \hlinewd{0.8pt}
    \end{tabular}
    }\vspace{-5pt}
   
\end{table}
\begin{figure}
	\centering
	\renewcommand{\thesubfigure}{}
	\subfigure[(a) CATs]
	{\includegraphics[width=0.495\linewidth]{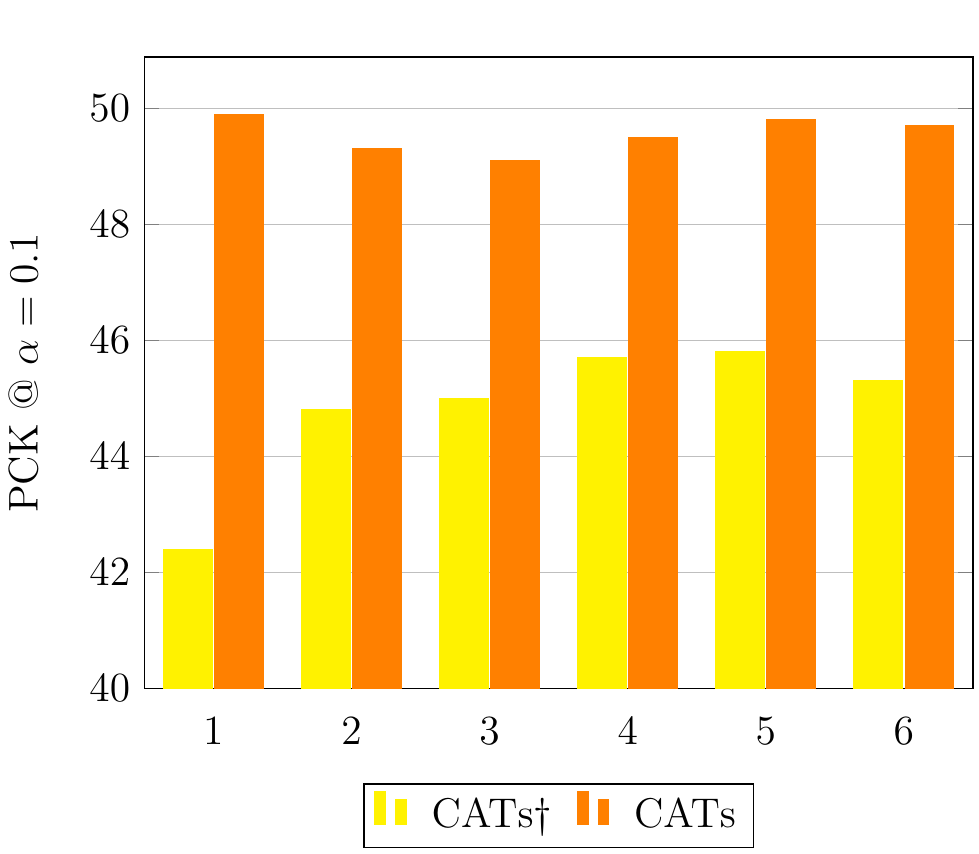}}\hfill
	\subfigure[(b) CATs++  ]
	{\includegraphics[width=0.495\linewidth]{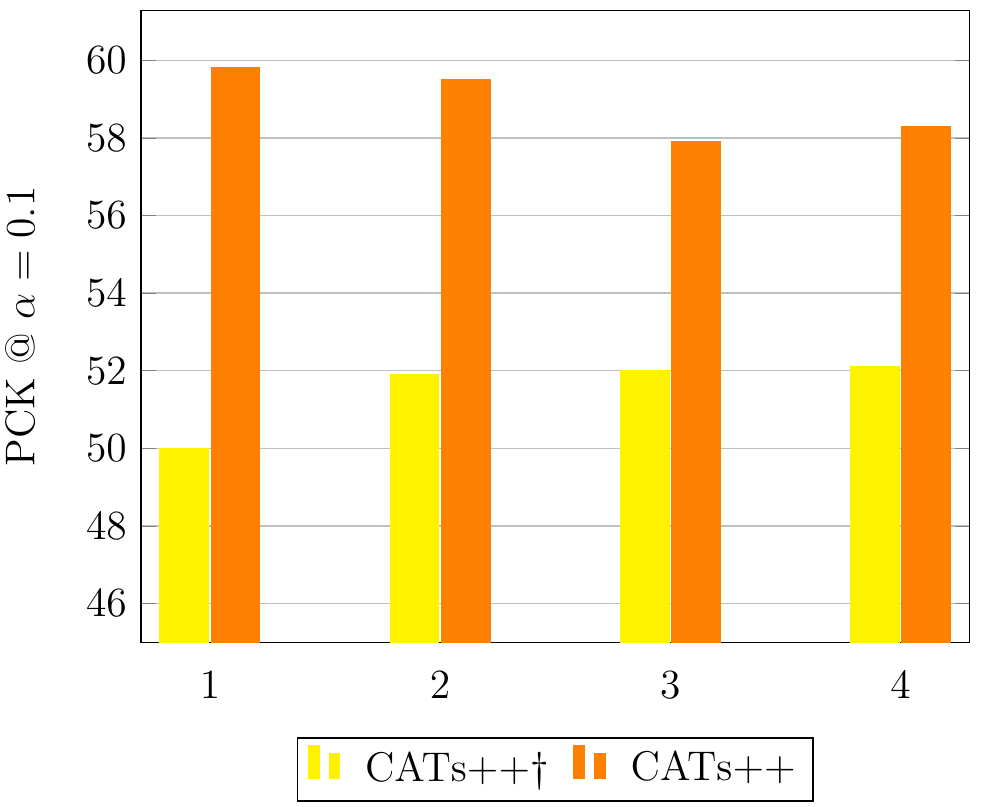}}\hfill\\\vspace{-10pt}
    \caption{\textbf{Effects of varying the number of encoders.} }
\vspace{-10pt}\label{depth}
\end{figure}

Summarizing the results, we consistently observe that leveraging multi-level features shows apparent improvements in performance, proving the effectiveness of multi-level aggregation. It is worth noting that DINO, which is more excel at dense tasks than DeiT-B, outperforms DeiT-B when applied to semantic matching. This indicates that the learned representation has a large influence on the final performance.  The interesting results we observed are the ones we obtained with CLIP-based backbone. Given ViT w/ CLIP$_\texttt{all}$ or ResNet-101 w/ CLIP$_\texttt{all}$ as backbone networks, we find that the performance is enhanced compared to the results with conventional ResNet-101. For ViT w/ CLIP$_\texttt{all}$ we were surprised to observe such a significant performance boost. Even for PF-PASCAL, the performance boost occurs, confirming that the use of CLIP backbone shows an apparent performance boost.

\subsubsection{Effects of varying the number of the encoders}
As done in numerous works~\cite{dosovitskiy2020image,carion2020end,li2020correspondence,sun2021loftr} that utilize transformers, we can also stack more encoders to increase their capacity and validate the effectiveness by varying the number of encoders. We show the effects of varying the number of encoders in Figure~\ref{depth}. It is natural that choosing a higher number of encoders inevitably increases the memory consumption and run-time, but it obtains a larger model capacity in return. This is confirmed by the results when the number of encoder is set to 1, where both CATs$\dagger$ and CATs++$\dagger$ report the lowest PCK. This indicates that the increase of the capacity of transformer aggregator clearly helps to boost the performance, which even surpasses the results reported in Table~\ref{tab:main_table}. Interestingly, showing completely contrary results, when the backbone networks are fine-tuned, the best PCKs are obtained when the number of encoders is set to 1. We suspect that trying to optimize both the backbone networks and the cost aggregator already risks overfitting, and by increasing the capacity of our cost aggregator, the model starts to overfit.

\begin{table}[]
    \centering
       \caption{\textbf{Effects of augmentation.}}
    \label{tab:aug}\vspace{-5pt}
    \begin{tabular}{l|c|cc}
\toprule
&\multirow{2}{*}{Augment.} &\multicolumn{2}{c}{SPair-71k}\\

& &\multicolumn{2}{c}{$\alpha_{\mathrm{bbox}}$ = 0.1}\\
               \midrule
DHPF~\cite{min2020learning} &\xmark &37.3&-\\
DHPF~\cite{min2020learning} &\cmark &39.4&-\\
\midrule
CHMNet~\cite{min2021efficient} &\xmark &-&47.0\\
CHMNet~\cite{min2021efficient} &\cmark &-&51.3\\
\midrule
CATs &\xmark & {35.1}&{43.5}\\
CATs &\cmark & {42.4}&{49.9}\\\midrule
CATs++ &\xmark & {\underline{47.3}}&\underline{54.0}\\
CATs++ &\cmark & {\textbf{50.0}}&\textbf{59.8}\\
\bottomrule
\end{tabular}\vspace{-5pt}
 
\end{table}

\subsection{Analysis}
\label{5.5}

\subsubsection{Data augmentation}
Transformer is well known for lacking some of the inductive bias and its data-hungry nature thus necessitates a large quantity of training data to be fed~\cite{vaswani2017attention,dosovitskiy2020image}. Recent methods~\cite{touvron2020deit,touvron2021going,liu2021swin} that employ transformers to address Computer Vision tasks have empirically shown that data augmentation techniques have a positive impact on performance. Also, to compensate for low generalization power caused by the provision of sparsely annotated data, \textit{i.e.,} keypoints, we provide a means to address it. Moreover, in the correspondence task, the question of to what extent can data augmentation affect performance has not yet been properly addressed. To this end, from the experiments, we empirically find that data augmentation has consistent positive impacts on performance in semantic correspondence.

In Table~\ref{tab:aug}, we compared the PCK performance between our variants, DHPF~\cite{min2020learning} and CHMNet~\cite{min2021efficient}. We note if the model is trained with augmentation. For a fair comparison, we evaluate all the methods trained on SPair-71k~\cite{min2019spair} using strong supervision, which assumes that the ground-truth keypoints are given. For CHMNet, we use the results reported in~\cite{min2021efficient}. The results show that compared to DHPF and CHMNet, two typical examples of CNN-based cost aggregation methods, data augmentation technique has a larger influence on CATs in terms of performance. This demonstrates that not only we ease the data-hunger problem inherent in transformers, but also found that applying augmentations for matching has positive effects. Specifically, this technique allowed the performance boosts for DHPF and CHMNet by 2.1$\%$p and 4.3$\%$p, respectively, but it is further boosted for CATs by 6.4$\%$p. This is further confirmed by the results of CATs++, showing that the augmentation technique brings apparent performance improvements by 5.8$\%$p.

\begin{table}[]
    \centering
    \caption{\textbf{Comparison of serial/parallel processing.}}
    \label{tab:serial}\vspace{-5pt}
    \begin{tabular}{l|cc|c}
\toprule
&\multirow{2}{*}{Serial} &\multirow{2}{*}{Parallel} &SPair-71k\\

&& &$\alpha_\mathrm{bbox}$ = 0.1\\
               \midrule

CATs$\dagger$ &\cmark&\xmark &\underline{42.4}\\
CATs$\dagger$ &\xmark &\cmark &38.3\\
CATs$\dagger$ &\cmark &\cmark &{\textbf{43.3}}\\\midrule
CATs  &\cmark&\xmark &\textbf{49.9}\\
CATs  &\xmark &\cmark &48.3\\
CATs  &\cmark &\cmark &{\underline{49.4}}\\\midrule
CATs++$\dagger$ &\cmark&\xmark &{45.5}\\
CATs++$\dagger$ &\xmark &\cmark &\textbf{50.0}\\
CATs++$\dagger$ &\cmark &\cmark &{\underline{46.6}}\\\midrule
CATs++  &\cmark&\xmark & 54.8\\
CATs++  &\xmark &\cmark &\textbf{59.8}\\
CATs++  &\cmark &\cmark &{\underline{55.9}}\\
\bottomrule
\end{tabular}\vspace{-5pt}
    
\end{table}

\begin{table}[]
\caption{\textbf{GPU memory and run-time comparison.} Inference time for aggregator is denoted by ($\cdot$) and subscript $coarse$ represents the coarsest layer.}
    \label{tab:memory}\vspace{-5pt}
\scalebox{0.9}{
    \centering
    \begin{tabular}{l|c|c|c}
\hlinewd{0.8pt}
&Aggregation &Memory [GB] &Run-time [ms]  \\
\midrule
NC-Net~\cite{rocco2018neighbourhood}& 4D Conv.&\textbf{1.2} &193.3\ (166.1)  \\
SCOT~\cite{liu2020semantic}&OT-RHM &4.6&146.5\ (81.6) \\
DHPF~\cite{min2020learning}&RHM &\underline{1.6} &57.7\ ({29.5})  \\
CHM~\cite{min2021convolutional}&6D Conv &\underline{1.6}&\underline{47.2}\ (38.3)\\
\midrule
CATs &Transformer &1.9&\textbf{34.5}\ (\underline{7.4})  \\
CATs++ &4D Conv. + Transformer &3.1&\ 110.2\ (60.6)  \\
CATs++$_{coarse}$ &4D Conv. + Transformer &\textbf{1.2}&57.4\ \ (\textbf{3.1})  \\
\hlinewd{0.8pt}
\end{tabular}%
}\vspace{-5pt}
    
\end{table}
\subsubsection{Serial/Parallel processing}
It is apparent that Equation~\ref{eq2} is not designed for an order-invariant output. Different from NC-Net~\cite{rocco2018neighbourhood}, we design CATs in a way that we let the correlation map undergo the self-attention module in a serial manner. We conducted a simple experiment to compare the difference between each approach. Moreover, we also provide a quantitative results comparison of CATs++ to justify our choice. For the experimental setting of serial processing, we sequentially aggregate the correlation map with a shared aggregator, while for parallel processing, we transpose the given input and let the original and transposed correlation maps undergo the same aggregator and add them subsequently.
\begin{figure}
	\centering
	\renewcommand{\thesubfigure}{}
	\subfigure[(a) ]
	{\includegraphics[width=0.475\linewidth]{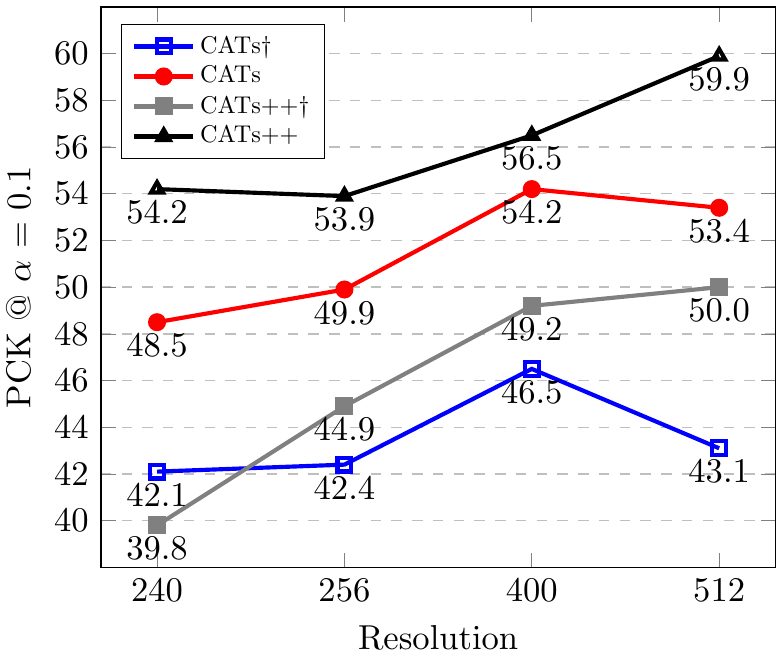}}\hfill
	\subfigure[(b) ]
	{\includegraphics[width=0.495\linewidth]{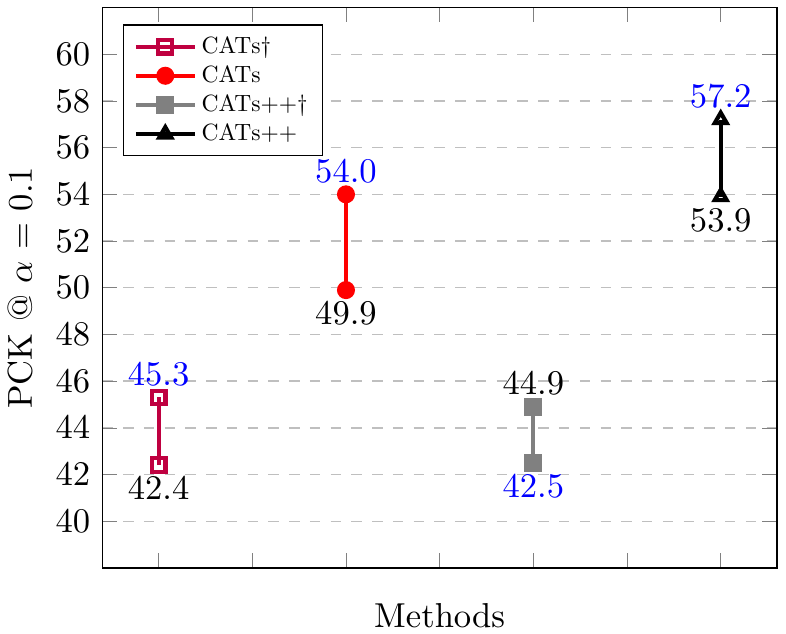}}\hfill\\\vspace{-10pt}
    \caption{\textbf{Ablation study of image resolution:} (a) as varying the image resolutions, (b) comparison between proposed methods at resolution set to 256. Note that we add results for CLIP-ResNet-101~\cite{radford2021learning}, which is written as blue color.}
\vspace{-10pt}\label{resolution}
\end{figure}
The results are summarized in Table~\ref{tab:serial}. We argue that although CATs may not support order invariance, adopting serial processing can obtain higher PCK as it has a better capability to reduce inconsistent matching scores by additionally processing the already processed cost map, which we finalize the architecture to include serial processing. However, interestingly, CATs++ clearly shows different results, which the parallel processing surpasses serial processing by a large margin. {Also, we observe that employing both serial and parallel processing does not yield higher performance than parallel processing. We conjecture that whether the network makes use of convolutions yields different results. More specifically, while token-mixing convolutional operations mix both spatial dimensions of correlation maps, self-attention only considers either source or target spatial dimension, and this makes self-attention suitable to serial processing and convolutions suitable to parallel processing. Note that through parallel processing, CATs++ aggregates local contexts of both spatial dimensions and the following self-attention impart them to all pixels, which allows the best results. We thus argue that employing serial processing for CATs++ may complicate the learning process by performing redundant aggregations, while it can benefit CATs. }

\subsubsection{GPU memory consumption and run-time comparison}
In Table~\ref{tab:memory}, we show the memory and run-time comparison to NC-Net~\cite{rocco2018neighbourhood}, SCOT~\cite{liu2020semantic}, DHPF~\cite{min2020learning}, and CHM~\cite{min2021convolutional}. For a fair comparison, the results are obtained using a single NVIDIA GeForce RTX 2080 Ti GPU and Intel Core i7-10700 CPU. We measure the GPU memory consumption given a single batch. We also measure the inference time for both the process without counting feature extraction and the whole process. Specifically, we note the type of aggregation method in the table, and the reported results represent the memory consumption and running-time of the aggregator, for instance, convolutional embedding module and efficient transformer aggregator for CATs++.  
Thanks to transformers' fast computation nature, compared to other methods, CATs show much faster inference time, allowing real-time inference. {This can benefit several applications which finding correspondences act as one of the milestones, \textit{e.g.,} object tracking and video object segmentation~\cite{perazzi2016benchmark}. Taking an example, an autonomous vehicles require real-time detection, tracking or segmentation of objects for a decision making, and a fast inference time can reduce delays during a decision making process, greatly reducing probability of potential incidents. Note that although CATs++ may suffer from relatively slower inference speed to other works~\cite{min2020learning,min2021convolutional}, less than 80 ms slower run time is a minor sacrifice for the better performance. }

For the memory consumption, we find that compared to other cost aggregation methods including 4D, 6D convolutions, OT-RHM, and RHM, the transformer aggregator shows comparable efficiency in terms of computational cost. Note that NC-Net and CHM show relatively lower memory consumption to others as they utilize a single feature map while all other methods utilize multi-level feature maps. Also, it is worth noting that CATs++ requires the largest memory consumption. This is because of the use of multi-level features as well as the processing of inputs at higher resolutions, which we show the reduced memory consumption and run-time at a coarse level of CATs++ to confirm this. Although CATs++ managed to reduce computational costs as summarized in Table~\ref{tab:memory}, processing the correlation maps at higher resolutions inherently requires much larger costs in exchange for large performance gain. Further enhancing the efficiency by balancing with the performance would be a promising direction, which we leave as future work. 

\subsubsection{Resolution of input images}
We find that different baseline methods~\cite{rocco2018neighbourhood,min2020learning,min2019hyperpixel,lee2019sfnet,liu2020semantic,min2021convolutional,cho2021semantic,min2021efficient,zhao2021multi,lee2021patchmatch,truong2021learning} for semantic correspondence does not use the fixed input resolutions at resolution according to the implementations authors provide. This means that comparing the proposed method to other baseline methods that use different resolutions may make the comparison unfair. We argue that to make a fair comparison, the training resolutions can be different among works, but the evaluation resolutions should be
the same.  To this end, we provide the results of our proposed methods at different resolutions to make fair comparisons to previous methods first, then we suggest a benchmark for semantic correspondence task for future works.

As shown in Fig.~\ref{resolution} (a), we observe that the higher resolution generally yields higher performance. Specifically, at resolution 400, CATs yields the best results, while at resolution 512, CATs++ yields the best results. This shows that the higher resolution generally helps to gain higher performance. However, it can be seen at resolution 512 that the performance of CATs drops severely. We suspect that this is the limitation of cost aggregation with a standard transformer, which its performance can not infinitely scale as the resolution increases. As seen in Fig.~\ref{resolution} (a), we argue that direct comparison is possible only when the results for different input sizes are given. Now taking this into account, we suggest a fixed benchmark for future works in semantic correspondence task.

\begin{figure}
    \centering
    \renewcommand{\thesubfigure}{}
    \subfigure[]
	{\includegraphics[width=0.495\linewidth]{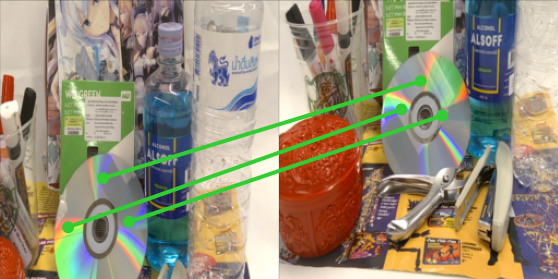}}\hfill
    \subfigure[]
	{\includegraphics[width=0.495\linewidth]{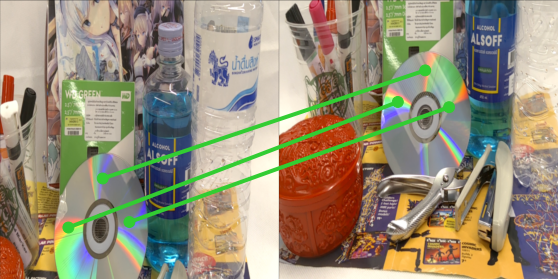}}\hfill\\
	\vspace{-20.5pt}
    \subfigure[]
	{\includegraphics[width=0.495\linewidth]{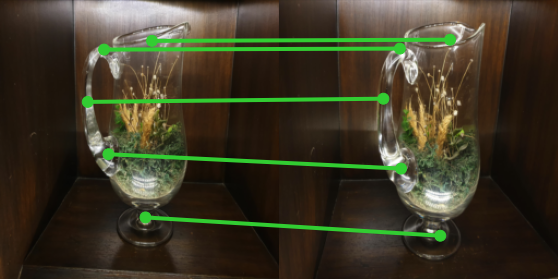}}\hfill
    \subfigure[]
	{\includegraphics[width=0.495\linewidth]{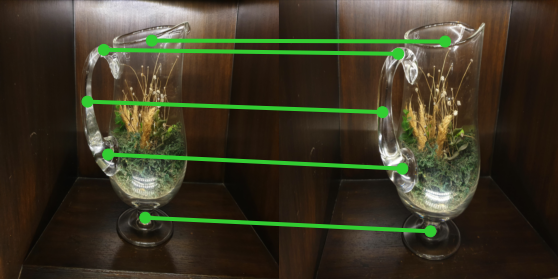}}\hfill\\
	\vspace{-20.5pt}
    \subfigure[(a) CATs]
	{\includegraphics[width=0.495\linewidth]{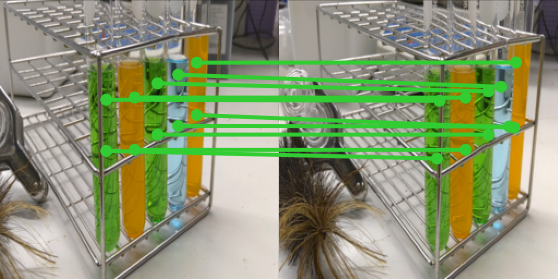}}\hfill
    \subfigure[(b) CATs++]
	{\includegraphics[width=0.495\linewidth]{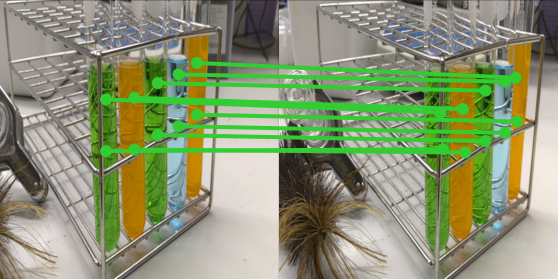}}\hfill\\
	\vspace{-10pt}
      \caption{\textbf{Qualitative results on non-lambertian objects.} The proposed methods can obtain accurate correspondence fields between non-lambertian objects.  }
\label{fig:teaser}\vspace{-10pt}
\end{figure}

We suggest the evaluation sizes be consistent across all the works for a fair comparison, which we decide as 256, and the results of the proposed methods at 256 evaluation resolution are shown in Fig.~\ref{resolution} (b). As shown, the performance of CATs++ dropped by a quite large margin. More specifically, we observe 6.0$\%p$ drop in performance for CATs++ when the evaluation size is changed from 512 to 256. This should make a fair comparison to other works evaluated at 256. Now we introduce a new finding to narrow this gap, which we adopt a new feature backbone, ResNet-101 initialized by pre-trained weights released by CLIP~\cite{radford2021learning}. We were surprised to observe that by simply replacing the conventional ResNet-101 pre-trained on ImageNet~\cite{deng2009imagenet} to that of CLIP, apparent performance boosts can be made. This can be a valuable finding that could bring an apparent performance boost even for other tasks, thanks to the improved training scheme and datasets by CLIP.

\begin{figure}
	\centering
	\renewcommand{\thesubfigure}{}
	\subfigure[(a) SPair-71k~\cite{min2019spair}]
	{\includegraphics[width=0.495\linewidth]{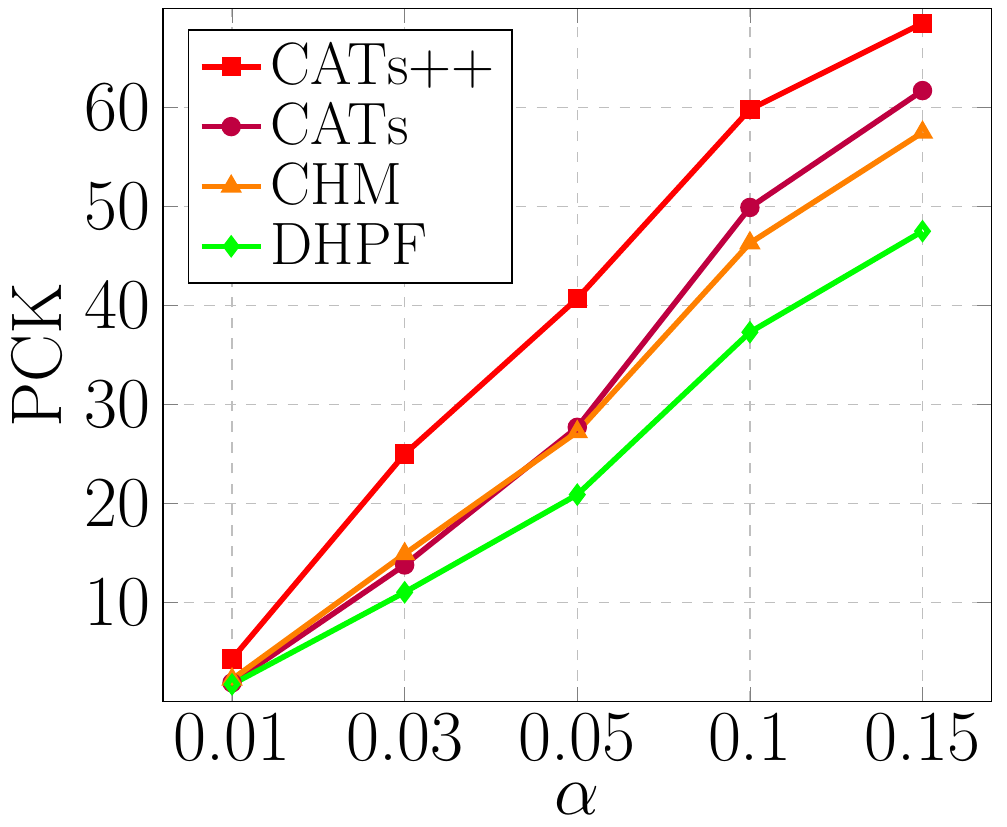}}\hfill
	\subfigure[(b) PF-PASCAL~\cite{ham2017proposal}]
	{\includegraphics[width=0.495\linewidth]{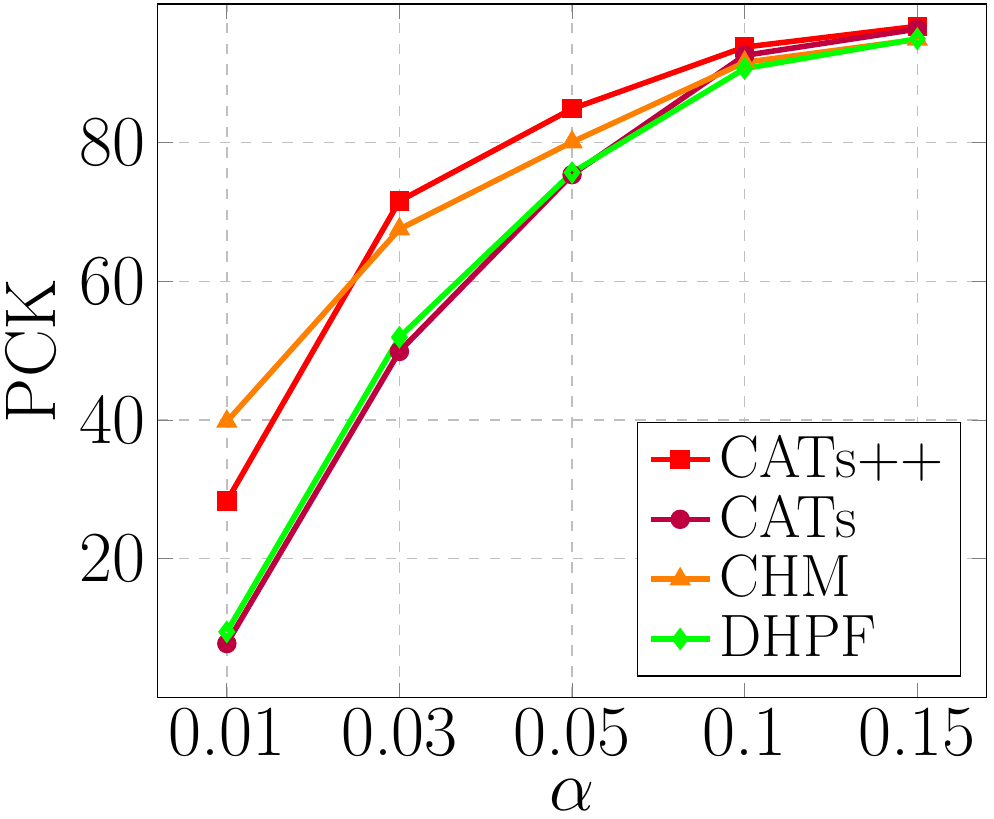}}\hfill\\
	\vspace{-10pt}
    \caption{\textbf{Quantitative results as varying the alpha threshold on SPair-71k~\cite{min2019spair} and PF-PASCAL~\cite{ham2017proposal}.} CATs++ captures fine-details surprisingly well compared to other methods (including CATs), thanks to its hierarchical architecture.}
\vspace{-10pt}\label{alpha}
\end{figure}
\subsubsection{Capturing fine details}
As an extension to CATs, we proposed CATs++, a novel cost aggregation approach that enjoys both reduced costs and boosted performance, and composed to have a hierarchical architecture, which as a result, it allowed the model to be excel at capturing fine-details as shown in Fig.~\ref{alpha}. As shown in Fig.~\ref{alpha}, where we evaluate DHPF~\cite{min2020learning}, CHM~\cite{min2021convolutional}, CATs and CATs++ on SPair-71k~\cite{min2019spair} and PF-PASCAL~\cite{ham2017proposal} by varying the alpha threshold used for the PCK evaluation, thanks to its hierarchical architecture, CATs++ successfully captures fine details, accurately finding the matching points, which CATs struggles to find. Note that CATs++ consistently outperforms all other methods except one case where CHM outperforms at $\alpha=0.01$ for PF-PASCAL. Interestingly, when the $\alpha$ threshold gets lower, CATs perform worse than other methods, indicating that it struggles to capture the fine details, which CATs++ successfully overcomes this limitation.

\subsubsection{Extension to few-shot segmentation}
We also show that the proposed methods perform well at a rudimentary level even for few-shot segmentation task. Few-shot segmentation aims to reduce reliance on labeled data and only a few support images with their associated masks are given for obtaining the segmentation maps for a query image. For this experiment, we simply replace soft-argmax with a 2D decoder as done in~\cite{min2021hypercorrelation} to produce a segmentation map and we show that CATs and CATs++ both also achieve competitive results on FSS-1000~\cite{li2020fss}. This is because few-shot segmentation shares a similar condition to semantic correspondence task, as it's also challenged by intra-class variations and background clutters. 

{In Table~\ref{tab:fss_sota}, we summarize the quantitative results on FSS-1000~\cite{li2020fss} dataset. FSS-1000~\cite{li2020fss}, a dataset specifically designed for few-shot segmentation, consists of 1000 object classes. Following~\cite{li2020fss}, 1000 classes are categorized into 3 splits for training, validation and testing, which consist of 520, 240 and 240 classes, respectively. For the evaluation metric, we employ mean intersection over union (mIoU), which averages over all IoU values for all object classes.   }

{We observed that CATs++ performs almost on par with other methods~\cite{wang2020few,liu2021few,min2021hypercorrelation} that are specifically designed for few-shot segmentation task. However, we find that the overall performance is generally lower than those of other works, especially when compared to  VAT~\cite{hong2022cost}. The reason for the performance gap can be explained by the difference in generalization power. As CATs only adopts Transformer as its aggregator, it may suffer from lower generalization power caused by the lack of inductive bias. While other works, including HSNet~\cite{min2021hypercorrelation}, VAT~\cite{hong2022cost} and CATs++, avoid this by utilizing convolutions for translation equivariance or adopting Swin Transformer~\cite{liu2021swin} that benefits from relative positioning bias. Although HSNet~\cite{min2021hypercorrelation} and CATs++ benefit from convolutional inductive bias just like VAT~\cite{hong2022cost} does, the best performance by VAT~\cite{hong2022cost} can be explained by the joint use of both convolutions and Swin Transformer~\cite{liu2021swin}, which allowed stronger generalization power compared to HSNet~\cite{min2021hypercorrelation} and CATs++ that only utilize convolutions within their architectures. Nevertheless, our approach showed that it is competent at both tasks despite its relatively lacking generalization power, and even outperforms other works in semantic matching. }

{Finally, comparing the memory consumption and the learnable parameters to HSNet~\cite{min2021hypercorrelation} and VAT~\cite{hong2022cost}, we observe that CATs requires much less memory consumption than all other approaches. It should also be noted that CATs++ is much more lightweight than VAT~\cite{hong2022cost}, which clearly demonstrates its efficiency. However, we observe that the number of learnable parameters for CATs and CATs++ are larger than other works, which is known to have an inverse relation to generalization power~\cite{Tetko1995NeuralNS}, which may explain the performance gap when the task is to predict segmenation maps of unseen classes. }

\begin{table}[]
    \centering
      \caption{\textbf{Quantitative comparison on FSS-1000~\cite{li2020fss}.}}\label{tab:fss_sota}
    \scalebox{0.9}{
        \begin{tabular}{clcccc}
                \toprule
                \multirow{2}{*}{\shortstack{Backbone\\feature}} & \multirow{2}{*}{Methods} & \multicolumn{2}{c}{mIoU} & {Memory}& {\# of learnable }\\ 
                
                & & 1-shot & 5-shot &[GB]&params.\\
                
                \midrule

                \multirow{4}{*}{ResNet50~\cite{he2016deep}}       &FSOT~\cite{liu2021few}&82.5&83.8&-&-\\
               &HSNet~\cite{min2021hypercorrelation} & \underline{85.5} & \underline{87.8}&-&2.6M\\
                & VAT~\cite{hong2022cost} &\textbf{90.1} &\textbf{90.7} &-&3.2M\\\cline{2-6}\\ [-2.0ex]
                & CATs++ &85.2 &85.4 &-&5.5M\\
                \midrule
                \multirow{5}{*}{ResNet101~\cite{he2016deep}} & DAN~\cite{wang2020few} & {85.2} & {88.1}&-&- \\ 
                
                & HSNet~\cite{min2021hypercorrelation}  & \underline{86.5} &\underline{88.5}&\underline{2.2}&\textbf{2.6M} \\ 
                & VAT~\cite{hong2022cost} &\textbf{90.3} & \textbf{90.8} &3.8&\underline{3.3M} \\\cline{2-6}\\ [-2.0ex]
                & CATs  &80.8 &80.9&\textbf{1.9}&4.6M \\
                & CATs++  &85.1 &85.3&{2.6}&5.5M \\
                
                \bottomrule
        \end{tabular}
        }
        \vspace{-2.0mm}
      \vspace{-10pt}
\end{table}

\begin{table}[b]
\centering

\resizebox{0.49\textwidth}{!}{%
\begin{tabular}{ll|ccc}
\toprule
   Method type &   Method        &  0.5m, 2\textdegree &  0.5m, 5\textdegree  & 5m, 10\textdegree  \\ \midrule
Sparse & D2-Net~\cite{dusmanu2019d2} & 	74.5 & 86.7 & \textbf{100.0} \\
& R2D2~\cite{revaud2019r2d2} & 69.4 & 86.7 & 94.9 \\ 
& R2D2~\cite{revaud2019r2d2} (K=20k) & 76.5 & \textbf{90.8} & \textbf{100.0} \\
& SuperPoint~\cite{detone2018superpoint} & 73.5 & 79.6 &  88.8 \\
& SuperPoint~\cite{detone2018superpoint} + SuperGlue~\cite{sarlin2020superglue} & \textbf{79.6} & \textbf{90.8} & \textbf{100.0} \\
& Patch2Pix & 78.6 & 88.8 & 99.0 \\ \midrule

Dense-to-sparse  & Sparse-NCNet~\cite{rocco2020efficient} & 76.5 & 84.7 & 98.0 \\
 & DualRC-Net~\cite{li2020dual} & \textbf{79.6} & \textbf{88.8} & \textbf{100.0} \\
 \midrule
Dense  & RANSAC-flow~\cite{shen2020ransac} + Superpoint~\cite{truong2021learning} & 74.5 & 87.8 & \textbf{100.0} \\
& PDC-Net~\cite{truong2021learning} & 76.5 & 85.7 & \textbf{100.0} \\
& PDC-Net~\cite{truong2021learning} + SuperPoint~\cite{detone2018superpoint} & \textbf{80.6} & 87.8 & \textbf{100.0} \\
& \textbf{CATs++ + SuperPoint~\cite{detone2018superpoint}} &62.2& 73.5& 93.9 \\
\bottomrule
\end{tabular}%
}
 \vspace{-2.0mm}
\caption{\textbf{Visual localization on the Aachen day-night dataset~\cite{sattler2018benchmarking}.}}\label{localization}
\end{table}

\subsubsection{Extension to Visual Localization}
Finally, we show that our approach also benefits the task of visual localization and evaluate on Aachen dataset~\cite{sattler2018benchmarking}. Visual localization aims to determine location from images by estimating the absolute 6 DoF pose of a query image with respect to the corresponding 3D scene model. To evaluate, we submit the results to the online evaluation benchmark~\cite{sattler2018benchmarking}.

Note that the proposed approach outputs a dense flow field, which makes it require non-trivial implementations and strategies  to apply to SfM 3D reconstruction without hurting the performance. To alleviate this issue, we employ SuperPoint~\cite{detone2018superpoint} to be responsible for the keypoint detection stage, and we then follow Hloc~\cite{sarlin2019coarse} outdoor localization pipeline to obtain the final localization results. To train our networks, we employ large-scale outdoor dataset, MegaDepth~\cite{MDLi18}. Given output flow maps from our networks, we up-sample the flow map to original image size and utilize keypoint coordinates already obtained from SuperPoint~\cite{detone2018superpoint} to find the corresponding keypoint at the other image. 

The results are summarized in Table~\ref{localization}. We observe that the proposed approach struggles to find accurate correspondences given high-resolution images, failing to localize with fine-details. Unlike other works, which include PDCNet~\cite{truong2021learning},  DualRC-Net~\cite{li2020dual} and SuperGlue~\cite{sarlin2020superglue}, our approach outputs the flow maps at relatively low resolution,~\textit{i.e.,} 32$\times$32. This prevents from capturing fine-details, an important aspect for the visual localization task.  Nevertheless, although a new state-of-the-art is not attained, a promising future direction is revealed, which is to capture fine details. As a future direction, a module that captures the missing details of the predicted flow can be designed. For example, local cropping around the coarse correspondence may be exploited to find fine details.

\begin{figure}
    \centering
    \renewcommand{\thesubfigure}{}
    \subfigure[]
	{\includegraphics[width=0.495\linewidth]{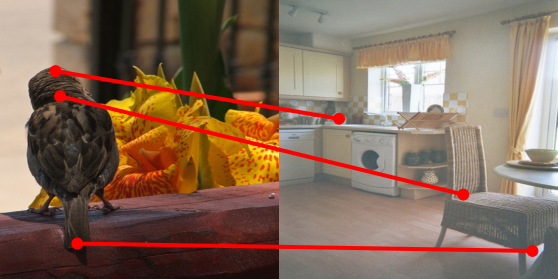}}
    \subfigure[]
	{\includegraphics[width=0.495\linewidth]{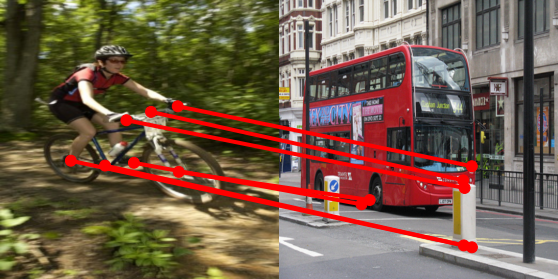}}\hfill\\
	\vspace{-20.5pt}
    \subfigure[]
	{\includegraphics[width=0.495\linewidth]{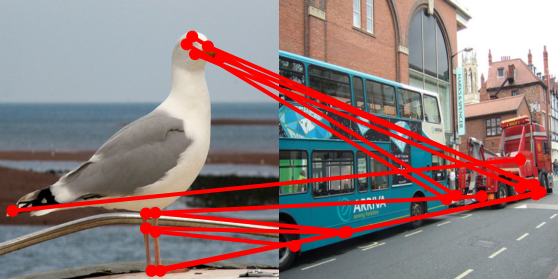}}
    \subfigure[]
	{\includegraphics[width=0.495\linewidth]{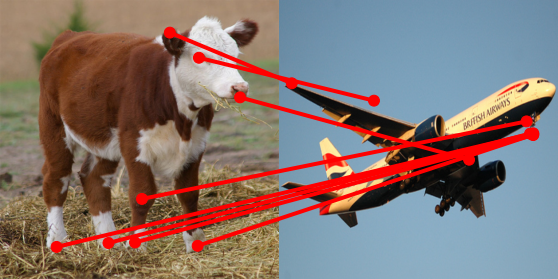}}\hfill\\
	\vspace{-20.5pt}
    \subfigure[]
	{\includegraphics[width=0.495\linewidth]{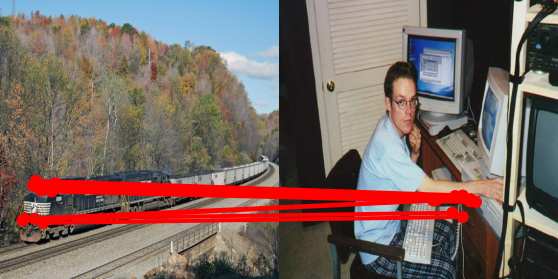}}
    \subfigure[]
	{\includegraphics[width=0.495\linewidth]{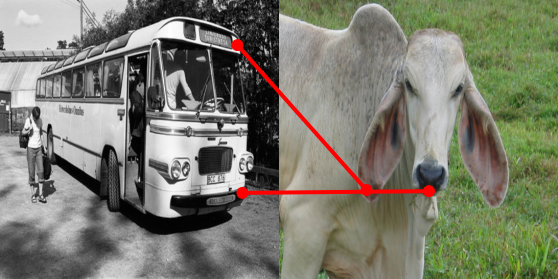}}\hfill\\
    \vspace{-15pt}
      \caption{\textbf{Failure cases.} Given images with irrelevant objects, the proposed method may struggle. }
\label{failure}\vspace{-10pt}
\end{figure}

\subsubsection{Limitations}
An apparent limitation of the proposed method is that it directly employs correlation maps, which are computationally expensive to process. This limits the resolutions of the input images the model can handle, enforcing the model to down-sample, then find the correspondences. Moreover, the proposed approaches assume that a pair of images with common objects, \textit{i.e.,} semantically similar, are given as inputs, making its applicability limited to only when the pair of images with semantically similar objects are given, as shown in Fig.~\ref{failure}.  Although such pairs can be obtained via image retrieval process, this may be one of the apparent limitations.

\section{Conclusion}

In this paper, we have proposed, for the first time, transformer-based cost aggregation networks for semantic correspondence which enables aggregating the matching scores computed between input features, dubbed CATs. We have made several architectural designs in the network architecture, including appearance affinity modelling, multi-level aggregation, swapping self-attention, and residual connection. We have shown that our method surpasses the current state-of-the-art in several benchmarks. Moreover, we extended CATs by introducing early convolutions and efficient transformer aggregator to reduce the computation costs and boost the performance, dubbed CATs++. We demonstrated the effectiveness of the proposed method on several benchmarks, which our method attains outperforms current state-of-the-art with large margin. We also conducted extensive ablation studies to validate our choices and explore its capacity and introduced expansion to other task, few-shot segmentation. 
\vspace{+10pt}

\noindent\textbf{Acknowledgements.}
This research was supported by the MSIT, Korea (IITP-2022-2020-0-01819, ICT Creative Consilience program), and National Research Foundation of Korea (NRF-2021R1C1C1006897). 

\ifCLASSOPTIONcompsoc
 
\else

  \section*{Acknowledgment}
\fi

\ifCLASSOPTIONcaptionsoff
  \newpage
\fi



\bibliographystyle{IEEEtran}
\bibliography{egbib}

%

\begin{IEEEbiography}[{\includegraphics[width=1in,height=1.25in,clip,keepaspectratio]{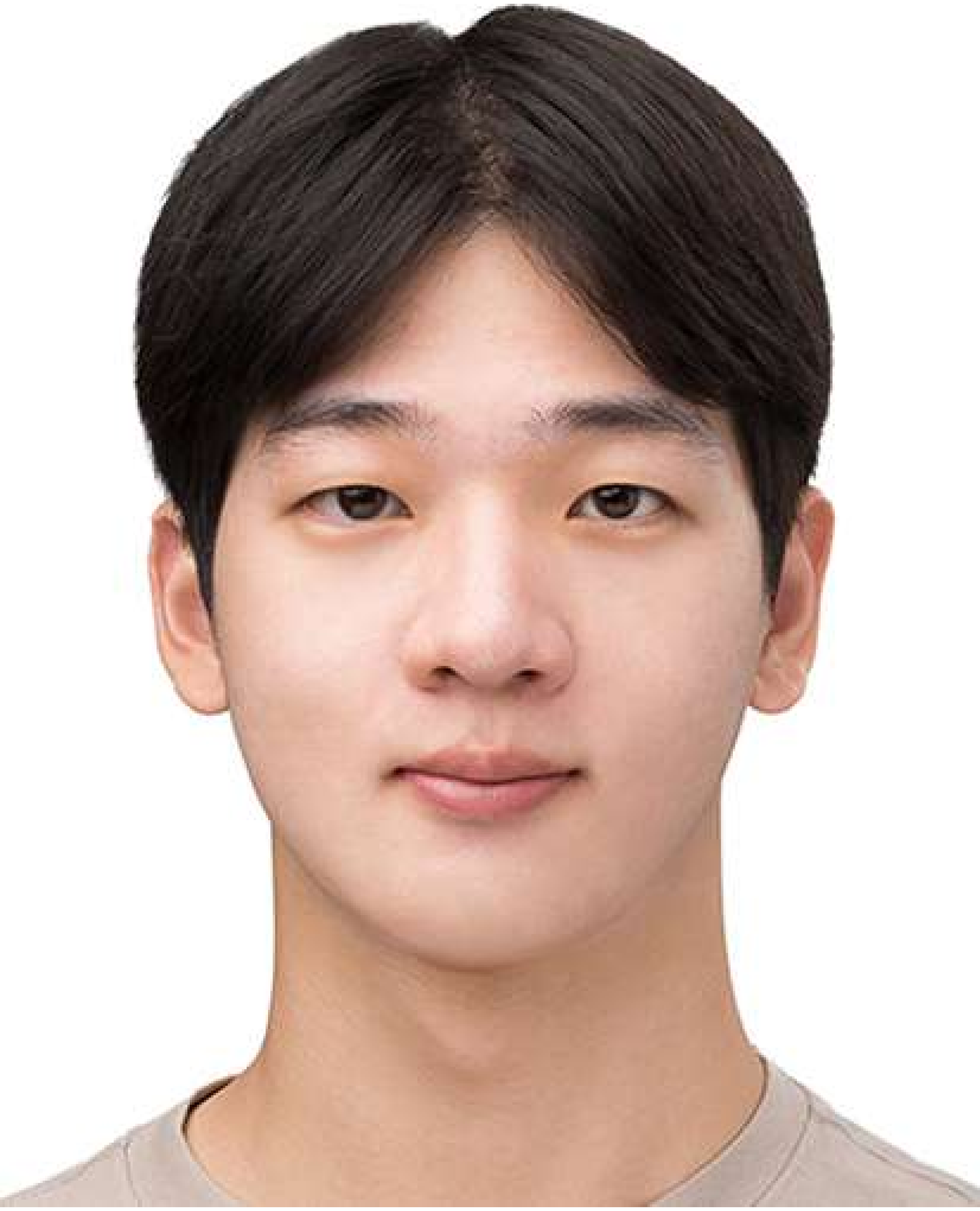}}]{Seokju Cho}
received his B.S. degree in Computer Science from Yonsei University in 2022, currently pursuing a Ph.D. degree at Korea University. His primary research interest is learning visual correspondences and its applications such as few-shot semantic segmentation and 3D reconstruction.
\end{IEEEbiography}

\begin{IEEEbiography}[{\includegraphics[width=1in,height=1.25in,clip,keepaspectratio]{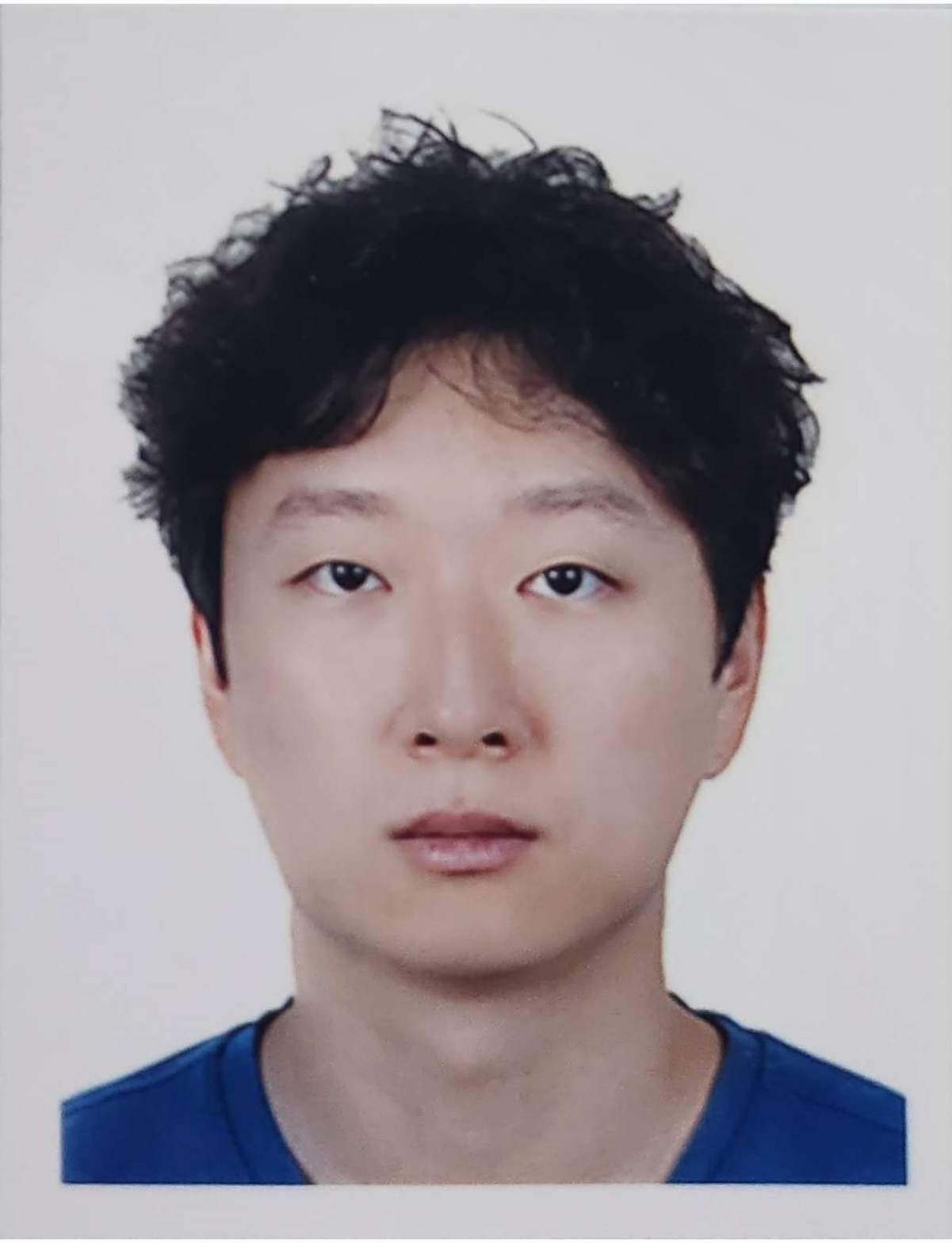}}]{Sunghwan Hong}
received his BS degree in Computer Sciences and Engineering from Korea University in 2021, where he is currently pursuing his PhD degree.
His current research focuses on visual correspondences and its applications such as few-shot learning and visual localization. 
He is also interested in 3D vision, multi-modal learning, NeRF and talking head generations.
\end{IEEEbiography}

\begin{IEEEbiography}[{\includegraphics[width=1in,height=1.25in,clip,keepaspectratio]{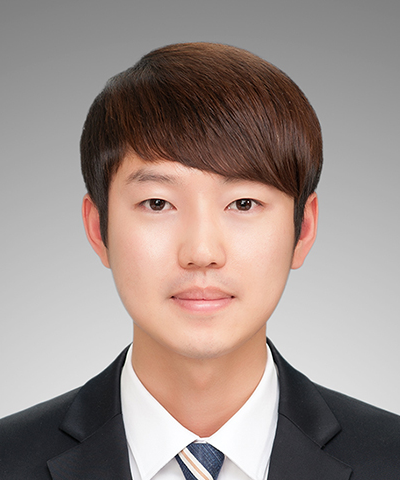}}]{Seungryong Kim}
received the B.S. and Ph.D. degrees from the School of Electrical and Electronic Engineering from Yonsei University, Seoul, Korea, in 2012 and 2018, respectively. From 2018 to 2019, he was Post-Doctoral Researcher in Yonsei University, Seoul, Korea. From 2019 to 2020, he has been Post-Doctoral Researcher in School of Computer and Communication Sciences at \'{E}cole Polytechnique F\'{e}d \'{e}rale de Lausanne (EPFL), Lausanne, Switzerland. Since 2020, he has been an assistant professor with the Department of Computer Science and Engineering, Korea University, Seoul. His current research interests include 2D/3D computer vision, computational photography, and machine learning.
\end{IEEEbiography}




\end{document}